\documentclass[10pt,twocolumn]{article}
\usepackage[letterpaper,margin=0.85in,columnsep=0.25in]{geometry}
\usepackage[round]{natbib}

\usepackage{booktabs,graphicx,array,xcolor,microtype,amsmath,amssymb,url}

\newcommand{\VxKevMinusJevPairedTV}{-0.033451}
\newcommand{\VxKevMinusJevPairedTVLower}{-0.045321}
\newcommand{\VxKevMinusJevPairedTVUpper}{-0.021883}
\newcommand{\VxJevComplementAboveOnePointFive}{0.000000}

\newcommand{\VxJevDisagreementAUC}{0.313949}

\newcommand{\VxKevComplementAboveOnePointFive}{0.673000}

\newcommand{\VxKevDisagreementAUC}{0.353411}

\newcommand{\VxBaseComplementAboveOnePointFive}{0.000000}

\newcommand{\VxBaseDisagreementAUC}{0.144522}

\newcommand{\VxPostComplementAboveOnePointFive}{1.000000}

\newcommand{\VxPostDisagreementAUC}{0.333159}

\title{Probability Contracts: Accuracy, Coherence, and Decisions Across LLM Interfaces}
\author{
Han Chen\\
Independent Researcher
\and
Yingrui Li\\
Independent Researcher
}
\date{}
\begin{document}
\maketitle
\begin{abstract}
Equivalent probability requests can lead to different decisions even when both reports are valid. We introduce \emph{probability contracts}, a benchmark that connects exact finite-world posteriors, validated event alignment, interface coherence, and failure-aware decision evaluation. Across four model--interface configurations on 1,000 worlds, Kev has lower aggregate canonical posterior error than Jev but larger complement and coarsening residuals; accuracy ordering varies by stratum. Jev's Event and Choice interfaces change the binary action on 32.8\% of valid pairs at defer cost 0.10. Post-hoc analyses show that disagreement certifies only 11--52\% of mean binary pair error and does not consistently outperform confidence for selection. An action-region characterization and a standard scoring-rule identity explain averaging's expected Brier guarantee relative to random interface selection, but not a decision-loss guarantee at each cost. The loss contrast takes both signs on a 99-cost grid for every configuration; small penalties where both policies beat deferral have pointwise intervals containing zero. Secondary checks specified before collection include a separate 400-root cohort, where Event/Choice effects remain configuration-dependent. A joint surface-order and answer-ID intervention shifts posttrained probabilities. Both bounded reasoning arms yield no valid probability reports, leaving their probability accuracy undefined. Probability contracts make these distinctions measurable by evaluating event semantics, posterior error, coverage, and decision cost together.

\end{abstract}

\section{Introduction}
A model can report an event's probability directly or through a distribution over outcomes. If the event and available information are unchanged, the reports share a mathematical target: the event's probability given the stated information. Their agreement is observable, but an outcome label alone cannot reveal the correct probability for an individual case. Here, the posterior is fixed by the stated world and evidence, rather than the model's beliefs.

We use finite worlds whose counts and evidence likelihoods determine an exact posterior. A \emph{probability contract} ties each report to its information, event, output mapping, and required relations, enabling joint evaluation of accuracy, coherence, and decision loss.

Jev is a hosted decision service with a binary Noul interface, called \emph{Event} here, and a categorical \emph{Choice} interface. For event $A$ with exact probability $220/317\approx0.694$, Jev reports 0.84 through Event and 0.97 through a two-option Choice request. A wrong answer costs 1 and deferral costs 0.10. Event leads to deferral; Choice leads to answering yes, with expected loss about 0.306. Both reports pass validation and concern the same event. Appendix~\ref{app:example} documents this post-hoc example; Section~\ref{sec:results} measures how often actions change.

For this example, the contract requires Event and binary Choice to report $q(A)$ under the same evidence, and $p(A)+p(\neg A)=1$. This separates three questions. \emph{Accuracy} measures error relative to the posterior. \emph{Coherence} concerns probability identities and agreement across equivalent requests. \emph{Decision value} concerns the loss of the action chosen from a report.

\paragraph{Contributions.} We introduce a benchmark combining exact targets, ten probability relations, and decision evaluation that includes failed responses. It quantifies how much error disagreement reveals and how much remains hidden. We also evaluate averaging and disagreement-based selection. For averaging, Proposition~1 and standard scoring-rule identities explain the role of decision cost.

The primary study compares four model--interface configurations: Jev-1.13.0, Kev's typed decision head, and Qwen3.5-4B Base and Posttrained with answer-token likelihood readouts. Kev assigns probabilities to supplied options. The appendix adds Mistral/OLMo readouts and secondary checks.

\paragraph{Terminology.} A \emph{world} describes four cells of a finite population. Its \emph{information set} is the evidence revealed by the request, such as a sensor reading. A \emph{root} groups a world, that evidence, and all related requests; world and root counts coincide here. The \emph{canonical} request asks for the full category distribution. Capitalized Event and Choice name interfaces; lowercase event $A$ names a set of outcomes. A \emph{stratum} is a generator--encoding pair; encodings use tables joined by cell identifier or nested cell records. \emph{Coverage} is the fraction of roots where the controller answers. The \emph{oracle action} minimizes loss under the exact posterior; \emph{headroom} is its loss reduction relative to always deferring at a given cost.

\section{Related Work}
\paragraph{Coherence and uncertainty evaluation.}
Proper-score dominance provides a classical justification for coherent forecasts \citep{predd2007coherence}. \citet{zhu2024incoherent} test identities and repeats; \citet{paleka2024consistency} measure consistency through related forecasts and arbitrage; and \citet{andrews2026dutch} study richer event relations and irrelevant context. \citet{andrews2026revealed} develops a general label-free rationality framework, while \citet{wagner2024contests} test equivalent span-probability factorizations. \citet{matta2026rethinking} separate structural coherence, faithfulness, and usefulness. Our benchmark measures these distinctions against exact targets: alongside disagreement, it quantifies the posterior error shared by reports and the loss of their induced decisions.

\paragraph{Known distributions and simulation.}
\citet{lovering2025numeric} find identity and order effects in prompt-implied numeric distributions. \citet{paruchuri2024odds} study reasoning about statistical distributions; \citet{lee2026forecastbench} evaluate simulated-world forecasts. Here, enumeration and event maps supply exact targets for transformed requests, enabling per-world error and decision-regret evaluation.

\paragraph{Calibration and readout.}
Calibration assesses reliability across cases sharing a forecast or confidence level \citep{vaicenavicius2019evaluating}; grouping loss can remain even after calibration \citep{perezlebel2023beyond}. Exact targets additionally let us measure each report's posterior error. We report both and fit no test-set recalibration \citep{guo2017calibration}. Readout and belief--choice sensitivity motivate distinguishing the reported quantity from the rule used to act on it \citep{mahaut2024confidence,kim2026protocol,yamin2026beliefs,kudum2026policies}. Label preferences are a possible readout confound \citep{zhao2021calibrate}, although a complement residual alone cannot identify their cause. Appendix~\ref{app:related} covers concurrent Jev studies.

\paragraph{Scoring rules and decisions.}
Proper scoring rules are related to threshold decisions \citep{schervish1989assessors,gneiting2007strictly}; Murphy diagrams compare their elementary components across thresholds \citep{ehm2016quantiles}. Our controller uses the reject option of \citet{chow1970reject}, and decision calibration connects calibration requirements to downstream actions \citep{zhao2021decisions}. We use this established theory to interpret the measured losses. \citet{zhang2026different} show that similar aggregate scores on ContractNLI can conceal different decisions across request conditions. Exact event posteriors additionally let us evaluate each action's expected loss and regret, alongside the error in the probability that selected it.

\section{Probability Outputs and Their Targets}
LLM interfaces expose several quantities as \emph{probability}: next-token probabilities, likelihoods of exact strings, and percentages generated as text. Each needs an explicit mapping to the events being evaluated.

Jev's Choice and Noul interfaces assign probabilities to the exhaustive events in the request \citep{typesafe2026api}; Kev's head assigns mass to supplied options \citep{kev2026release}. For likelihood readouts, we take the model's next-token probabilities for the answer IDs and renormalize them to sum to one. We also record the total probability those IDs received before normalization (Appendix~\ref{app:diagnostics}). A separate confidence field is an event probability only if its documented meaning supports that interpretation.

Validation checks the answer set, finite nonnegative values, and total mass within tolerance. Valid reports are then scored for posterior error, coherence, and decision loss.

\section{Probability Contracts}
\subsection{Measurement object}
A probability contract specifies the event, available evidence, output mapping, and probability relations to be checked. Equivalent requests preserve the event and evidence; complement and conditional requests are linked by probability identities. Requests and responses record the evidence used to evaluate the contract.

Each world has four latent cells indexed by Boolean attributes $a,b$. The request supplies each cell's attributes, category, integer prior count $w_i$, and rational evidence likelihood $\ell_i$, plus the sampling rule. Counts determine prior sampling weights; $\ell_i$ gives the probability of the observed evidence, or a compatibility indicator for an observed attribute. The posterior mass is
\[
q_i=\frac{w_i\ell_i}{\sum_jw_j\ell_j}.
\]
Probabilities of events and categories sum their cells' posterior masses. The two encodings supply the same kinds of information: \emph{joined tables} separates attributes, counts, and likelihoods into three tables linked by cell identifier; \emph{nested records} groups them within each cell. Appendix~\ref{app:example} shows a complete stored request.

In the introductory example, counts $(67,85,9,79)$ and likelihoods $(1/2,9/10,1,1/2)$ give weights $(33.5,76.5,9,39.5)$. The first two categories in label order form event $A$, so $q(A)=110/158.5=220/317\approx0.694$.

Four generators vary the evidence and categories. \emph{Finite-population} conditions counted objects on $b$ when observed. \emph{Noisy-sensor} supplies a sensor reading's likelihood for each cell. \emph{Joint-attributes} samples counted cells with no additional evidence. \emph{Missing-rule} conditions on $a$ when observed and groups cells into three categories. The others have four categories.

The exact-posterior calculator uses rational arithmetic. A separately implemented path normalizes the prior in 70-digit decimal arithmetic before the likelihood update; admission requires agreement within $10^{-15}$. A typed event algebra checks atoms, set membership, negation, conjunction, disjunction, and conditioning. Partitions must be disjoint and exhaustive on positive posterior support, and conditioning events must have positive mass.

\subsection{Representations and relations}
Each root has a canonical request and eleven variants. The mathematical event $A$ contains the first $\lfloor K/2\rfloor$ sorted categories; $\neg A$ is its complement. The event $B$ is $b=\mathrm{True}$, with requests for $A\cap B$ and $A\mid B$. Event and binary Choice ask about the same event $A$; an exact repeat duplicates Event. Other variants permute options, change layout within an encoding, replace option keys with category-uninformative identifiers (opaque keys), or add independently assigned metadata.

Each option maps to a set of cells: its \emph{event map}. Aligning reports means matching their coordinates to these same sets before comparison. A \emph{residual} measures deviation from a relation satisfied by the exact posterior. Table~\ref{tab:relations} separates probability identities, representation changes, and exact-repeat variability. Residuals are reported separately because the tests measure different discrepancies.

\begin{table*}[t]
\caption{Declared probability relations after event alignment. Coarsening compares an event report with the sum of its categories in the canonical report. Fr\'echet bounds require $\max(0,p(A)+p(B)-1)\le p(A\cap B)\le\min(p(A),p(B))$. Continuous residuals are primary; pass thresholds are prespecified summaries.}
\label{tab:relations}\centering\small
\resizebox{\textwidth}{!}{\begin{tabular}{llll}
\toprule
Relation & Aligned claims & Residual or condition & Eligibility\\
\midrule
Complement & $p(A),p(\neg A)$ & $|p(A)+p(\neg A)-1|$ & Exhaustive binary event\\
Coarsening & $p(A),p(\omega)$ & $|p(A)-\sum_{\omega\in A}p(\omega)|$ & Same partition and information\\
Event/Choice & $p_{\rm event}(A),p_{\rm choice}(A)$ & Absolute difference & Bijection to same event\\
Intersection & $p(A),p(B),p(A\cap B)$ & Distance outside Fr\'echet bounds & Aligned $A,B$\\
Product & $p(A\cap B),p(B),p(A\mid B)$ & $|p(A\cap B)-p(B)p(A\mid B)|$ & $q(B)>0$\\
Permutation & $p(\omega),p_\pi(\omega)$ & Total variation after inverse map & Bijection; same information\\
Rendering & $p(\omega),p_r(\omega)$ & Total variation & Same encoding family\\
Opaque keys & $p(\omega),p_k(\omega)$ & Total variation after key map & Bijection\\
Irrelevant metadata & $p(\omega),p_m(\omega)$ & Total variation & Metadata independent by design\\
Exact repeat & $p(A),p'(A)$ & Absolute difference & Identical payload\\
\bottomrule
\end{tabular}}
\end{table*}

Transformations retain the world, evidence, target, and option-to-event map. Vector comparisons invert this map; scalar comparisons use the event algebra. Undefined conditionals are never sent. Product comparisons are eligible on 913 of 1,000 roots; all other variants are eligible on every root. Prespecified tolerances summarize pass rates.

\section{Experimental Design}
\subsection{Worlds and configurations}
The primary cohort contains 1,000 roots: 250 per generator and 125 per generator--encoding stratum, using joined tables or nested records. Semantic hashes separate it from 2,700 earlier benchmark roots; they do not establish absence from pretraining data. Twelve requests per root give 12,000 intended slots per configuration. Excluding 87 undefined conditionals leaves 11,913 eligible requests.

The local configurations are Kev-4B, Qwen3.5-4B-Base, and posttrained Qwen3.5-4B \citep{kev2026checkpoint,qwen2026base,qwen2026post}. Kev's adapter and pointer head assign probabilities to the supplied options \citep{kev2026release}. Its published training and temperature-calibration details \citep{kev2026training}, pinned checkpoints, and stored temperature appear in Appendix~\ref{app:kev-provenance}. No calibration is fitted here. All three configurations use FP32; Posttrained uses its non-thinking template. Kev's rounded output is a separate sensitivity. We use Kev, Base, Posttrained, and Jev as short labels, with precision-qualified names for secondary configurations.

Jev-1.13.0 is queried through its native Noul and Choice interfaces. A separate 477-request development set checks technical compatibility and is excluded from estimates. Prompts, thresholds, and the main cohort were fixed before collection. Each main request has one recorded attempt. Jev returns serialized decimal probabilities, making numerical precision part of the evaluated configuration.

A supplementary check uses Mistral and OLMo with likelihood and unconstrained generated-JSON readouts on 200 reused worlds (Appendix~\ref{app:extension}). Appendix~\ref{app:prospective-stress} reports the reasoning comparison, joint recoding, and the \emph{headroom cohort}: 400 roots selected by binary posterior confidence to increase the scope for beneficial class actions.

\subsection{Validity and scoring populations}
Collection cannot access evaluator posteriors. Exact requests and raw responses are recorded before validation, including rejected responses. Each eligible request is valid, rejected after a response, or unavailable. Probability metrics use valid outputs. Paired-valid comparisons require both reports to be valid on the same root; multi-report relations require every constituent. Decision evaluation includes every intended root, assigning rejected or unavailable outputs to defer. Low-cost fallback can reduce loss by reducing coverage, so we report validity, coverage, and loss together.

Native validity is the valid fraction of eligible requests. Service availability additionally counts rejected HTTP-200 responses as served. Bounded identical-payload retries are permitted only for designated transient failures; no main Jev request used that allowance (Appendix~\ref{app:collection}).

The operational criterion fixed before Jev collection requires native validity at least 0.995 and a lower 95\% root-bootstrap pass-rate bound at least 0.95 for each of the nine primary relations in Table~\ref{tab:relations}, excluding its intersection-bounds diagnostic. Repeat uses tolerance 0.02; the others use 0.05. Continuous residuals accompany this threshold-based criterion.

\subsection{Accuracy, decisions, and uncertainty}
Posterior error is primarily $\operatorname{TV}(p,q)=\tfrac12\lVert p-q\rVert_1$. Expected multiclass Brier is $\sum_yq_y\sum_k(p_k-\mathbf 1[k=y])^2$, and expected log loss is $-\sum_yq_y\log p_y$, using reported probabilities without smoothing. Positive $q_y$ with $p_y=0$ gives infinite log loss, which remains in the result. Realized scores use one outcome draw per root. Prespecified calibration compares top-class confidence with these draws using fixed 10-, 15-, and 20-bin schemes and an equal-frequency sensitivity. A separately labeled post-hoc summary instead uses the exact posterior mass of the selected class. Both describe grouped reliability; TV measures per-root error. No recalibration is fitted.

The reject-option controller \citep{chow1970reject} chooses a class or defers. Wrong and correct answers cost 1 and 0; deferral costs $c$. The primary cost is 0.10, with 0.02, 0.05, 0.20, and 0.40 also fixed before collection. The controller minimizes expected cost under the report. For a normalized vector it acts when $\max_i p_i\ge1-c$, otherwise deferring; ties prefer the first canonical class before defer. Expected loss evaluates the chosen action under $q$, and oracle regret subtracts the minimum loss achievable under $q$. Always defer is the cost-$c$ reference; a posterior-informed best fixed action is descriptive only.

Canonical multiclass decisions and binary event decisions are scored separately. For a binary report $p(A)$, the controller uses $(1-p(A),p(A))$. Action-change rates compare aligned binary Event/Choice requests on paired-valid roots. Coverage accompanies action-change rates: two policies that always defer have identical actions even when their probabilities differ.

\begin{figure*}[t]
\centering
\includegraphics[width=\textwidth]{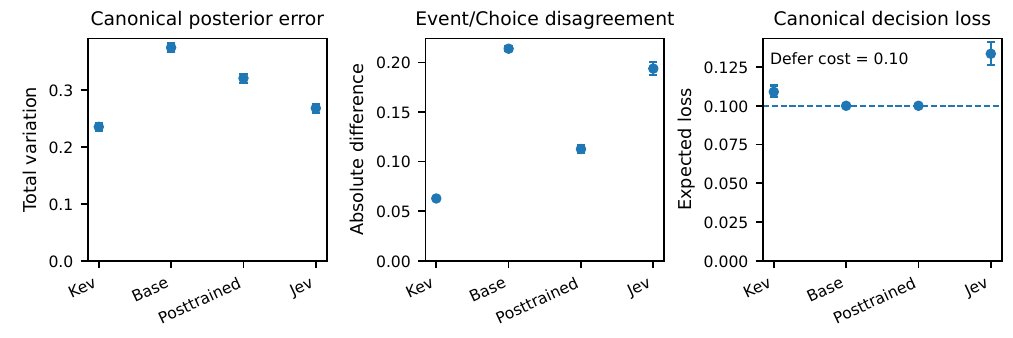}
\caption{Canonical posterior error (left), binary Event/Choice disagreement (middle), and canonical decision loss (right); lower is better. Error bars are 95\% stratified root-bootstrap intervals. Jev TV uses 990 valid canonical roots; other estimates use 1,000 roots. The outer panels concern the canonical multiclass task; the middle compares reports of the same binary event. The dashed line marks always-defer loss.}
\label{fig:main-summary}
\end{figure*}

Aggregates weight the eight generator--encoding strata equally. Valid-only estimates average their nonempty strata and disclose omissions. We use 5,000 paired root-bootstrap replicates (seed 2026092408), resampling within strata with all related requests and configurations together. Each statistic retains its own validity denominator. Intervals are pointwise percentile summaries over worlds. They provide neither multiplicity adjustment nor uncertainty over training runs or repeated service collection.

\paragraph{Analysis status.} Primary targets, relations, costs, and comparisons were specified before Jev collection. Applying the Jev criterion to all configurations (Appendix~\ref{app:symmetric}), exact-posterior calibration, rounding and outage-window sensitivities (Appendix~\ref{app:continuity}), and Section~7/Appendices~\ref{app:posthoc}--\ref{app:accelerated-offline} are post hoc. The disagreement analysis followed inspection of its descriptive ratios. Appendix~\ref{app:prospective-stress} collections were specified after primary results but before their own outputs. Appendix~\ref{app:reasoning-budget} adds a later diagnostic and post-collection scoring. These statuses apply throughout.

\section{Primary Study: 1,000 Worlds}\label{sec:results}
Figure~\ref{fig:main-summary} compares posterior error, interface disagreement, and decision loss.

\subsection{Posterior accuracy}
Kev has the lowest aggregate canonical TV, followed by Jev, Posttrained, and Base (Appendix Table~\ref{tab:accuracy}). On the 990 roots valid for both Kev and Jev, the paired Jev-minus-Kev difference is 0.0335 [0.0222, 0.0448]. Appendix~\ref{app:fulltables} gives all six paired contrasts, with each comparison's support.

The aggregate ranking varies across strata. Kev and Jev each have lower TV in four of the eight generator--encoding combinations. Kev's advantage is concentrated in the joint-attributes and noisy-sensor strata; Jev performs better in the missing-rule strata (Appendix Table~\ref{tab:strata}).

Jev assigns zero mass to possible outcomes in 20 of its 990 valid canonical reports, giving infinite expected log loss. Its five infinite realized losses concern the sampled outcomes. All original local canonical log losses are finite. Exact-posterior calibration and the prespecified sampled-outcome summaries are reported separately in Appendix~\ref{app:symmetric}.

\subsection{Coherence and representation stability}
Accuracy and coherence rank the configurations differently (Table~\ref{tab:coherence}). Kev's complement residual is 0.5055 versus Jev's 0.0929; coarsening residuals are 0.3554 and 0.1751. Conversely, Jev's Event/Choice gap exceeds Kev's and Posttrained's. Posttrained has the largest complement residual, 0.8000, and Kev the largest intersection-bound residual, 0.1553. A post-hoc signed-sum check finds excess mass in every Kev and Posttrained complement pair (Appendix~\ref{app:diagnostics}). The cause is unresolved.

\begin{table*}[t]
\caption{Mean relation residuals (lower is better within a row). Jev uses 1,000 valid pairs for Event/Choice, complement, and repeat; 990 for coarsening; 999 for intersection; 912 for product; and 976 for permutation. Local support is complete (913 eligible product roots, otherwise 1,000). All intervals and relations are in Appendix~\ref{app:allrelations}.}
\label{tab:coherence}\centering\small
\begin{tabular}{lrrrr}
\toprule
Relation & Kev & Base & Posttrained & Jev\\
\midrule
Event/Choice & 0.0629 & 0.2138 & 0.1126 & 0.1938\\
Complement & 0.5055 & 0.1155 & 0.8000 & 0.0929\\
Coarsening & 0.3554 & 0.2451 & 0.2836 & 0.1751\\
Intersection bound & 0.1553 & 0.0814 & 0.0358 & 0.0200\\
Product identity & 0.1328 & 0.2682 & 0.1406 & 0.0836\\
Permutation TV & 0.1143 & 0.3224 & 0.3470 & 0.0534\\
Exact repeat & 0.0000 & 0.0000 & 0.0000 & 0.0188\\
\bottomrule
\end{tabular}
\end{table*}

Joint recoding swaps the displayed no/yes option order and their assigned answer IDs. It changes Posttrained expected Brier by $-0.1971$ [$-0.2075$, $-0.1871$] on 2,000 event/complement reports. A post-hoc check gives complement residual 0.4404 versus 0.8000 before recoding (Appendix~\ref{app:prospective-stress}). The two-coordinate score contrast measures the combined intervention.

Exact repeats provide a distinct control. Local deterministic repeats agree exactly, whereas Jev's mean repeat residual is 0.0188. Although this mean lies below the 0.02 tolerance, only 745/1,000 pairs pass: 0.7450 [0.7180, 0.7720]. The mean residual and pairwise pass rate thus summarize different aspects of repeatability. Variability includes the served system as well as the model.

Jev fails the operational criterion: validity is below 0.995, and all nine primary relation pass-rate lower bounds are below 0.95. Applying the same criterion to the local configurations also gives no pass; this symmetric extension is post-hoc (Appendix~\ref{app:symmetric}). The individual residuals show how the failures differ. Appendix~\ref{app:allrelations} includes rendering, opaque keys, and irrelevant metadata, with eligibility, support, and intervals.

\subsection{Decision consequences}
Table~\ref{tab:decision} separates canonical multiclass loss, regret, and deferral from binary Event/Choice action changes. Thus Posttrained can always defer on the canonical task yet change actions between binary requests.

For binary Event/Choice at $c=0.10$, actions change on 14.9\% of roots for Kev, 34.2\% for Posttrained, and 32.8\% for Jev. Base always defers and has no action changes despite its probability gap. All 328 Jev changes are act/defer crossings. At $c=0.40$, 62 of 405 changes instead switch class labels. These action changes occur when valid reports of the same event cross a decision boundary; exact posteriors let us evaluate the resulting loss.

\begin{table*}[t]
\caption{Decisions at defer cost 0.10: means and 95\% root-bootstrap intervals. The first three metrics use the canonical multiclass task; the final column compares separate binary Event/Choice requests. Each uses 1,000 roots per configuration. Canonical loss includes failure-triggered deferral. Always-defer loss is 0.10.}
\label{tab:decision}\centering\small
\begin{tabular}{lrrrr}
\toprule
& \multicolumn{3}{c}{Canonical multiclass task} & Binary task\\
\cmidrule(lr){2-4}\cmidrule(lr){5-5}
Configuration & Expected loss & Oracle regret & Defer rate & Event/Choice changes\\
\midrule
Kev & \shortstack{0.1090\\{}[0.1054, 0.1131]} & \shortstack{0.0193\\{}[0.0158, 0.0232]} & \shortstack{0.9610\\{}[0.9500, 0.9720]} & \shortstack{0.1490\\{}[0.1290, 0.1710]}\\[5pt]
Base & \shortstack{0.1000\\{}[0.1000, 0.1000]} & \shortstack{0.0103\\{}[0.0088, 0.0118]} & \shortstack{1.0000\\{}[1.0000, 1.0000]} & \shortstack{0.0000\\{}[0.0000, 0.0000]}\\[5pt]
Posttrained & \shortstack{0.1000\\{}[0.1000, 0.1000]} & \shortstack{0.0103\\{}[0.0088, 0.0118]} & \shortstack{1.0000\\{}[1.0000, 1.0000]} & \shortstack{0.3420\\{}[0.3210, 0.3630]}\\[5pt]
Jev-1.13.0 & \shortstack{0.1336\\{}[0.1263, 0.1414]} & \shortstack{0.0438\\{}[0.0366, 0.0515]} & \shortstack{0.8160\\{}[0.7920, 0.8380]} & \shortstack{0.3280\\{}[0.3020, 0.3540]}\\
\bottomrule
\end{tabular}
\end{table*}

On the canonical task at $c=0.10$, oracle loss is 0.0897: only 0.0103 average improvement is available below always defer, with positive headroom on 119/1,000 roots. Kev defers on 96.1\% of roots yet incurs loss 0.1090, exceeding always defer by 0.0090. Jev's loss is 0.1336; Base and Posttrained equal always defer. At $c=0.40$, Kev's loss is 0.3906, now 0.0094 below always defer. The five-cost grid is shown in Appendix Figure~\ref{fig:decision-cost-grid}.

For this loss, posterior TV $\delta$ bounds oracle regret by $2\delta$, or by $\delta$ when the predicted and oracle actions differ only between acting and deferring. This guarantee uses posterior error; agreement between reports alone is insufficient. Appendix~\ref{app:geometry} gives the elementary bounds and the margins needed to preserve an action across representations.

\section{What Disagreement Misses and Averaging Changes}
The post-hoc analyses in this section quantify error hidden by agreement and evaluate two ways of using paired reports: averaging and agreement-based deferral. Each configuration has 1,000 valid Event/Choice pairs and their repeats. Policies use only the reports and declared cost; exact posteriors are used afterward to evaluate their actions.

The target here is binary event $A$. At $c=0.10$, binary oracle loss is 0.08132, leaving 0.01868 available below always defer, with positive headroom on 208 roots.

\subsection{Disagreement bounds only part of error}
Let $p,p'$ report the same event with exact probability $q$, and let $m=(p+p')/2$. Two standard identities give
\begin{align}
\frac{|p-q|+|p'-q|}{2}
 &= \frac{|p-p'|}{2}+d(q,[p\wedge p',p\vee p']),\label{eq:absolute-certificate}\\
\frac{(p-q)^2+(p'-q)^2}{2}
 &= \frac{(p-p')^2}{4}+(m-q)^2.\label{eq:squared-certificate}
\end{align}
Here $d$ is the distance from $q$ to the interval between the reports. Half their absolute gap, the \emph{disagreement lower bound} (what disagreement certifies), is observable without $q$. Exact targets reveal the remaining error. The squared identity separates disagreement from the error of the midpoint.

The disagreement lower bound accounts for 11--52\% of mean pair error across configurations (Appendix~\ref{app:posthoc}). On these same binary pairs, Jev also has lower mean absolute error than Kev (0.18567 versus 0.29397), despite its larger Event/Choice gap (0.1938 versus 0.0629). Thus the accuracy--coherence reversal persists when both measures use the same event and paired roots. When the mean report supplies class predictions on selected worlds, retaining low-disagreement pairs at matched 50\% coverage gives lower point-estimated loss than random retention for Jev and Posttrained. It beats confidence selection only for Jev. All these 50\%-coverage policies exceed always-defer loss at $c=0.10$ (Appendix~\ref{app:accelerated-offline}). These selection results support a configuration-specific diagnostic, not a general improvement in decisions. Appendix~\ref{app:posthoc} gives absolute components, intervals, and complement checks.

\subsection{A better Brier score need not mean a better decision}
Averaging is the equal-weight Euclidean projection of $(p,p')$ onto equality. By Equation~\ref{eq:squared-certificate}, replacing the pair with $m$ reduces its average expected one-coordinate Bernoulli Brier score by $(p-p')^2/4\ge0$. The outcome-variance term $q(1-q)$ cancels; the two-coordinate categorical binary convention doubles all terms. The comparator in this guarantee \citep{predd2007coherence} is \emph{Random interface}: choose Event or Choice with equal probability, independently of the world.

The connection to decisions is an average over costs. Writing $\ell_c(x;q)$ for the binary controller's expected loss, Appendix~\ref{app:scoring-bridge} proves, with the stated boundary ties,
\[
2\int_0^{1/2}\ell_c(x;q)\,dc=(x-q)^2+q(1-q).
\]
The expected Brier score is therefore the average loss under a uniform distribution over these costs. A score improvement can coexist with higher loss at an individual cost.

\paragraph{Proposition 1 (averaging and action regions).}
For $0<c<1/2$, let $a_c(x)$ be no for $x\le c$, defer for $c<x<1-c$, and yes for $x\ge1-c$. Their true losses are $q,c,1-q$. Writing $\ell_c(x;q)=L_c(a_c(x);q)$, the averaging penalty is
\begin{equation}
\Delta_c=\ell_c(m;q)-\tfrac12\{\ell_c(p;q)+\ell_c(p';q)\}.
\label{eq:averaging-penalty}
\end{equation}
It is zero if both reports select the same action. For adjacent action regions it equals half the selected-minus-unselected true action loss. For opposite class actions it is $q-1/2$, $c-1/2$, or $1/2-q$, according as the midpoint selects no, defer, or yes. In particular, resolving opposite class actions to defer always helps. The proof is a case analysis on convex action regions (Appendix~\ref{app:averaging-geometry}). Thus gap size alone cannot determine the sign: the cost, crossed regions, and true posterior matter.

We compare Event and Choice with two fixed two-query policies: use the mean report (\emph{Mean}), or take the common action and otherwise defer (\emph{Agreement}). Event plus its exact repeat supplies equal-query controls for both rules. Missing constituents would trigger defer. Random interface requires one query; Mean and Agreement require two.

\begin{table*}[t]
\caption{Post-hoc binary policy loss at $c=0.10$ (all 1,000 roots per configuration; lower is better). Parentheses in headers give requests per decision. Agreement means common action or defer. Always-defer loss is 0.10000 and binary oracle loss is 0.08132. Intervals and full comparisons are in Appendix~\ref{app:posthoc}.}
\label{tab:binary-policies}\centering\small
\begin{tabular}{lrrrrr}\toprule
Configuration & Event (1) & Choice (1) & Mean (2) & Agreement (2) & Repeat mean (2)\\\midrule
Kev & 0.16695 & 0.15786 & 0.15281 & 0.14186 & 0.16695\\
Base & 0.10000 & 0.10000 & 0.10000 & 0.10000 & 0.10000\\
Posttrained & 0.25448 & 0.23257 & 0.19055 & 0.17239 & 0.25448\\
Jev & 0.09654 & 0.17052 & 0.09890 & 0.09654 & 0.09645\\
\bottomrule\end{tabular}\end{table*}

At the primary cost $c=0.10$ (Table~\ref{tab:binary-policies}), Mean reduces loss relative to Random interface by 0.03463 for Jev, 0.00959 for Kev, and 0.05297 for Posttrained. Base stays at always defer. Kev and Posttrained remain worse than always defer under both two-interface policies.

A separate comparison asks whether averaging beats a fixed individual interface. For Jev at $c=0.10$, averaging increases loss over Event by 0.00236 [0.00077, 0.00411], while coverage rises from 0.062 to 0.104. Both losses remain below always defer. The Brier guarantee compares Mean with Random interface; this contrast compares Mean with Event alone. Agreement equals Event loss; the Event/repeat mean has loss 0.09645 at 0.057 coverage.

\begin{figure}[t]
\centering
\includegraphics[width=\columnwidth]{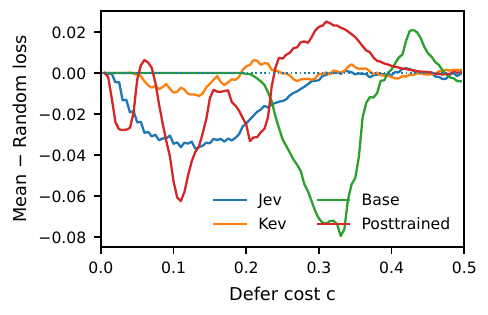}
\caption{Mean minus Random interface expected binary loss at the 99 evaluated costs (post hoc). Positive values favor Random interface. Lines connect grid estimates; Appendix~\ref{app:accelerated-offline} supplies uncertainty and exact breakpoint records.}
\label{fig:main-cost-contrast}
\end{figure}

On the original five-cost grid, averaging has higher mean loss than the same Random interface used by the Brier guarantee in 3 of 20 configuration--cost cells: Kev at $c=0.20$, and Base and Posttrained at $c=0.40$ (Appendix Table~\ref{tab:binary-cost-grid}). In all three, both policies are worse than always defer. Posttrained's penalty is 0.00347 [0.00122, 0.00581]. The post-hoc 99-cost sweep (Figure~\ref{fig:main-cost-contrast}) changes the scope: the contrast takes both signs for every configuration, including small penalties where both policies beat defer. Every such penalty on that grid has a pointwise interval containing zero (Appendix~\ref{app:accelerated-offline}). The intervals leave population-level harm unresolved. Appendix~\ref{app:averaging-geometry} constructs arbitrarily accurate reports for which both policies beat deferral; the model experiments leave that regime unresolved.

\section{Discussion and Limitations}
Knowing each world's posterior lets us separate disagreement from shared error and score decisions exactly. These per-world quantities cannot be inferred from one outcome label alone, which motivates controlled finite worlds.

The practical implication is to evaluate probability interfaces with their decision rules. Low residuals can coexist with substantial error; averaging can improve a proper score while increasing loss at a particular cost. Validity and coverage matter because cheap fallback deferral can make a failing interface appear competitive on loss alone.

Event/Choice Brier contrasts change sign across configurations in the headroom cohort, and joint recoding reduces the Posttrained complement residual (Appendix~\ref{app:prospective-stress}). The original 160-token (bounded) reasoning arms yielded no valid reports. The later exploratory diagnostic compares 12--15 roots per request type where both modes returned valid reports in the final collection stage (Appendix~\ref{app:reasoning-budget}, Table~\ref{tab:reasoning-exploratory}). Thinking lowers mean TV and Brier for all five types, but raises loss at $c=0.10$ and lowers it at $c=0.40$. All valid thinking reports classify at the lower cost; controls mostly defer. This small, incomplete sample leaves population effects unresolved.

Four-cell worlds and two encodings permit exact targets; generalization to open-ended tasks remains untested. The primary local models share a Qwen lineage, and hosted results cover one Jev revision. The 7B extension has only 59 and 60 valid canonical JSON reports out of 200, and its identical Event/Choice inputs are identity controls. Training, prompts, readout, and precision vary across configurations. The comparisons therefore do not isolate a single component; substantial posterior errors still leave high-accuracy behavior unresolved.

The headroom cohort is a selected population. Limited primary-cohort headroom restricts possible decision gains. The post-hoc analyses identified in Section~5 are exploratory; their pointwise intervals give no continuum-wide guarantee. Historical service order and omitted monetary query costs limit policy comparisons. The appendix and artifact document the evidence and replay limits.

\section{Conclusion}
Probability contracts tie each report to a defined event, information set, and exact posterior target. On the evaluated configurations, agreement hides shared error, equivalent requests change actions, and averaging has cost-dependent decision effects despite its Brier guarantee. Evaluating accuracy, coherence, and decision loss together reveals these differences; validity and coverage keep failures visible.

\clearpage
\section*{AI Use Statement}
The human authors contributed substantially to the study's conceptual development and experimental design. Generative AI tools supported parts of this work, including refinement of hypotheses, mathematical claims and proofs, methodology, implementation, data processing, analysis, and interpretation. ChatGPT and Codex assisted with benchmark and analysis code, testing, experiment orchestration, literature discovery, figures, artifact preparation, and manuscript drafting and editing. Claude suggested or refined the action-region and cost-mixture arguments and secondary analyses involving cost curves, disagreement-based selection, joint recoding, the headroom cohort, and reasoning. Claude and Gemini also provided methodological and manuscript critiques.

AI-assisted code generated the synthetic finite-world instances and exact posterior targets and processed collected outputs. Instances were generated deterministically from the specified design. Evaluated model responses remained as collected; writing assistants supplied no replacement predictions or reference posteriors. Translation, qualitative or thematic analysis, surveys, interviews, and transcription were not components of this study.

Verification used separately implemented rational and high-precision decimal posterior calculators, exact-rational checks of mathematical identities and boundary cases, semantic tests separating report-based action selection from target-based scoring, independent checks of exported calculations, and source checks for citations. The artifact identifies checks that can be rerun from the supplied data and collection-stage checks documented by summaries. The authors take responsibility for the final claims, interpretations, text, and artifacts. Generative AI systems are not authors.

\bibliographystyle{apalike}
\bibliography{references}

\clearpage
\appendix
\onecolumn
\section{Secondary Interface and Headroom Checks}\label{app:prospective-stress}

These secondary checks were specified after the primary results and before their own collection. The reasoning comparison reuses the original 1,000 roots; the hosted and local checks use the 400-root headroom cohort. The joint recoding compares recoded requests with historical reports. Results remain separate from the primary estimates. Every component used zero retries, and evaluator targets were withheld from collection. All outputs were retained as collected. Probability metrics use valid reports and their applicable support; the reasoning and headroom decision metrics assign rejected or missing reports to the predeclared defer action. The joint recoding comparison reports probability changes and policy consequences on the same retained reports.

\paragraph{Scoring and uncertainty.}
For the reasoning comparison and headroom-cohort checks, accuracy compares the report's maximizing class with the sampled label, not the posterior mode. Expected Brier uses the full categorical vector, including both coordinates for binary reports. The sampled-label accuracy is descriptive; TV remains the direct posterior-error measure. These checks use the separately specified costs $c\in\{0.05,0.10,0.20,0.30,0.40\}$, distinct from the primary grid; results are not pooled across grids. Their 95\% intervals use 5,000 stratified percentile bootstrap resamples of roots within generator--encoding strata. The reasoning comparison and Jev use seed 20260923; the local headroom configurations use deterministic metric- and configuration-specific seeds derived from 2026092804. The intervals describe individual contrasts without a multiple-comparison adjustment.

\paragraph{FP32 and BF16 non-thinking configurations.}
The primary Posttrained readout uses FP32. The reasoning comparison uses the same checkpoint and original roots with a BF16 non-thinking control. The control's canonical TV, Event/Choice gap, and complement residual are 0.4129, 0.2341, and 0.3921; the primary values are 0.3202, 0.1126, and 0.8000. Complete input equivalence was not verified, so these are separate configurations and the differences cannot be attributed to precision alone. The artifact lists the available configuration records.

\paragraph{Reasoning comparison on the original cohort.}
Across the original 1,000 roots and five variants, the two Qwen3.5-4B arms completed 10,000 requests with recorded outcomes. The BF16 non-thinking control produced 5,000/5,000 valid reports, including 1,000 canonical reports and 1,000 Event/Choice pairs. Canonical accuracy was 0.306, mean TV 0.4129, and expected Brier 0.8090. The Reasoning arm produced 0/5,000 valid reports: all requests ended at the specified 160-token reasoning limit. With no valid reports, probability accuracy and the reasoning-versus-control accuracy contrast are undefined; the controller defers on every request. Appendix~\ref{app:reasoning-budget} reports the recovered development ledger and runtime-based cap selection, together with a later completion diagnostic and its separately specified exploratory scoring. In the valid control, Event/Choice absolute residual was 0.2341, complement residual was 0.3921, and repeat residual was 0.0000. Choice minus Event expected Brier was 0.1900 [0.1745, 0.2054], and Choice had higher point-estimated expected decision loss at all five specified costs.

\paragraph{Joint recoding.}
The Posttrained control changes the historical binary surface order from \texttt{no, yes} to \texttt{yes, no} while reassigning answer IDs. Both outputs are mapped back to the same events before comparison. All 2,000/2,000 reports were valid. Mean signed $P(\mathrm{yes})$ change was $-0.1882$ [$-0.1913$, $-0.1849$], mean absolute change was 0.2139 [0.2109, 0.2170], and swapped minus historical expected Brier was $-0.1971$ [$-0.2075$, $-0.1871$]. The 95\% intervals quantify the combined effect of surface order and answer-ID reassignment; the individual effects are not identified. The contrast averages event and complement reports, unlike the reasoning control's Choice-minus-Event contrast on event $A$ alone. Both are two-coordinate Brier differences. Squared posterior distance has the same paired difference because the target-dependent outcome-variance term cancels.

The variant means are $-0.0550$ for event $A$ and $-0.3392$ for its complement: most of the aggregate score reduction comes from complement prompts. A post-hoc check of these same 1,000 paired roots gives mean complement residual 0.4404 after recoding versus 0.8000 before. The residual after recoding still indicates a complement violation. No calibration is fitted.

\paragraph{Policy consequences of the joint recoding.}
Actions are selected from the report and cost only; evaluator targets are used afterward to score the fixed action. Table~\ref{tab:label-policy-corrected} retains all five original costs. Expected Brier includes outcome variance, which cancels in the paired common-target contrast. Root-paired intervals use 5,000 stratified resamples and metric-key-derived random streams with base seed 2026092803; the artifact supplies the seed mapping and a scoring-correction note. The loss difference is positive at $c=0.02$ and negative at the other four costs, so the Brier reduction is not a decision-loss guarantee.

\begin{table*}[t]
\centering\small
\caption{Policy consequences of the joint surface-order and answer-ID intervention. Both event $A$ and its complement are retained (2,000 paired reports on 1,000 roots); the two variant differences are averaged within root before stratified resampling. Loss differences are recoded minus historical, and action disagreement compares the two report-based controllers. Intervals are pointwise 95\% summaries. All policies use the original reports, with action selection independent of evaluator truth.}
\label{tab:label-policy-corrected}
\begin{tabular}{lrrrr}
\toprule
Defer cost & Action disagreement & 95\% interval & Loss difference & 95\% interval\\
\midrule
0.02 & 0.0535 & [0.0495, 0.0575] & $0.0240$ & [$ 0.0217$, $ 0.0263$]\\
0.05 & 0.1980 & [0.1865, 0.2095] & $-0.0083$ & [$ -0.0147$, $ -0.0020$]\\
0.10 & 0.6085 & [0.5980, 0.6195] & $-0.1917$ & [$ -0.2006$, $ -0.1827$]\\
0.20 & 0.6780 & [0.6660, 0.6905] & $-0.2020$ & [$ -0.2093$, $ -0.1943$]\\
0.40 & 0.2705 & [0.2615, 0.2795] & $-0.0341$ & [$ -0.0411$, $ -0.0270$]\\
\bottomrule
\end{tabular}
\end{table*}

\paragraph{Headroom cohort and reproducibility.}
The headroom cohort has 50 roots in each of eight generator--encoding strata. The pre-collection plan specifies boundary $[0.5,0.75)$, medium $[0.75,0.95)$, and strong $[0.95,1)$ bins, with quotas of 10, 10, and 30 roots per stratum and $q(A)$ endpoints excluded. Applying these thresholds to the retained binary oracle confidence $\max\{q(A),1-q(A)\}$ reproduces all 400 recorded bin labels and all eight strata's quotas. The supplied cases satisfy these criteria; the original builder is unavailable for exact sampler regeneration. At $c=0.10$, recorded mean oracle headroom is 0.04021 for the binary event and 0.01234 for the canonical task. These quotas give the headroom cohort a distribution different from the primary cohort.

\paragraph{Hosted headroom cohort.}
Jev-1.13.0 completed 2,000 requests with recorded outcomes on the headroom cohort: 1,992 were valid and eight ended as simplex-validation failures. All eight rejected records were canonical, leaving 392/400 valid canonical reports; each binary variant had 400/400 valid reports. Event/Choice contrasts use all 400 paired roots. Rejections remain excluded from probability metrics and trigger deferral in decisions on all intended roots. Event/Choice absolute residual was 0.1342, complement residual was 0.0844, and repeat residual was 0.0182. Choice minus Event expected Brier was 0.0645 [0.0463, 0.0841]. Choice had higher expected decision loss at all five costs, with differences from 0.0127 at $c=0.05$ to 0.0513 at $c=0.40$; all five intervals excluded zero. The expected-Brier difference is positive in all eight stratum means. Decision-loss differences instead have mixed signs across strata at some costs; the aggregate decision effects are not uniform stratum-level effects.

\paragraph{Local headroom cohort.}
The local checks completed 8,000 requests with recorded outcomes, 2,000 per configuration: 6,000 valid reports and 2,000 trace-cap rejections. Table~\ref{tab:h-local-stress} reports all configurations. The BF16 Reasoning arm produced 0/2,000 valid reports: every request reached the same specified 160-token limit. The valid configurations had deterministic repeat residual zero but different interface and complement residuals. On the same binary target, Choice minus Event expected Brier was $-0.0185$ [$-0.0325$, $-0.0048$] for Kev, 0.0537 [0.0222, 0.0854] for Base, and $-0.0253$ [$-0.0516$, 0.0007] for Posttrained. The Posttrained interval includes zero, and the favored interface differs across configurations. At $c=0.10$, the corresponding action-disagreement rates were 0.1725, 0, and 0.3400. The model--interface configuration, scoring target, and operating cost remain distinct parts of the comparison.

\begin{table*}[t]
\caption{Local results on the 400-root headroom cohort. Valid counts requests across variants. Accuracy uses the sampled label; accuracy, TV, and expected Brier each use 400 valid canonical reports per non-reasoning configuration. Gap denotes mean absolute Event/Choice difference; all three relation residuals use 400 valid pairs per non-reasoning configuration. Dashes indicate undefined probability metrics for Reasoning: all 2,000 requests reached the reasoning limit. }
\label{tab:h-local-stress}\centering\small
\resizebox{\textwidth}{!}{%
\begin{tabular}{lrrrrrrr}
\toprule
Configuration & Valid & Accuracy & TV & Exp. Brier & Gap & Complement & Repeat\\
\midrule
Kev (FP32) & 2,000/2,000 & 0.460 & 0.3426 & 0.6167 & 0.0772 & 0.5217 & 0.0000\\
Base (FP32) & 2,000/2,000 & 0.365 & 0.4326 & 0.7185 & 0.2109 & 0.1186 & 0.0000\\
Posttrained (FP32) & 2,000/2,000 & 0.420 & 0.4001 & 0.6786 & 0.1156 & 0.7929 & 0.0000\\
Reasoning (BF16) & 0/2,000 & -- & -- & -- & -- & -- & --\\
\bottomrule
\end{tabular}}
\end{table*}

\paragraph{Interpretation.}
The headroom-cohort checks supply new-cohort evidence; the reasoning comparison reuses the primary roots, and joint recoding compares against historical reports. The valid configurations show different interface sensitivities. The original two Reasoning arms fail to produce probability reports within their stated limits, leaving their accuracy unmeasured. The evidence concerns these finite-world configurations and their specified decision costs.

\begin{figure*}[t]
\centering
\includegraphics[width=\textwidth]{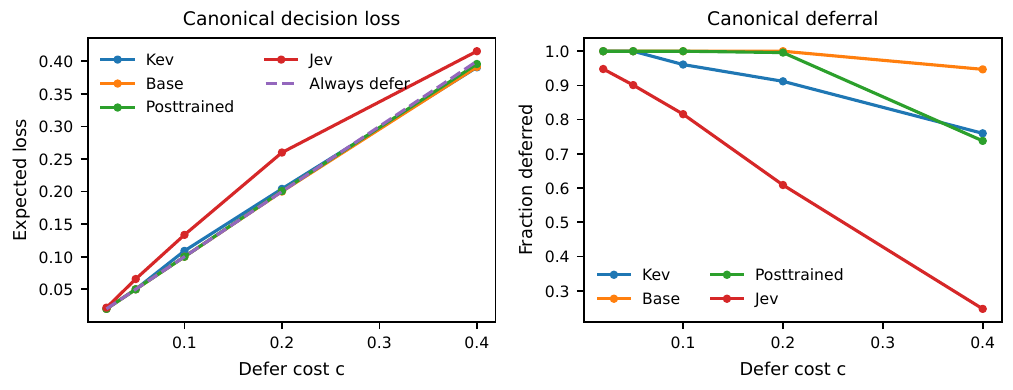}
\caption{The five prespecified costs, connected to show the sampled trend. Expected decision loss and deferral are shown separately; the dashed loss line is always defer. The primary cost is 0.10. Discussed in Section~\ref{sec:results}.}
\label{fig:decision-cost-grid}
\end{figure*}
\section{Model Training Provenance}
\subsection{Documented Kev training provenance}\label{app:kev-provenance}
The evaluated checkpoints are Kev-4B (\texttt{485ace8...}), Qwen3.5-4B-Base (\texttt{1001bb4...}), and posttrained Qwen3.5-4B (\texttt{851bf6e...}). Kev adds a rank-16 LoRA adapter and a 256-dimensional pointer head. We retain its stored temperature $T=2.143546925$, fitted on in-distribution development data, without a study-specific fit.

The release card at pinned source revision \texttt{557598f} describes the \texttt{decision-v7} suite: 10,000 public classification records, 896 policy minimal-pair records, and 1,680 generated-rule records. The LoRA adapter and newly initialized pointer head were trained on option-distribution cross-entropy. A later update mixed 1,425 generated date/evidence-removal records with 2,000 replay records. The stored temperature was fitted by negative log-likelihood on in-distribution development rows \citep{kev2026training}. These are release-author descriptions; the compact artifact does not independently bind the training manifests to the evaluated checkpoint. The benchmark's semantic-hash separation from its own 2,700 earlier roots is not a training-data overlap audit. We have not established whether the released training suites contain benchmark-equivalent worlds.

\section{Full Variant Inventory}
Table~\ref{tab:variant-inventory} lists each request slot. The undefined conditional slots remain in the inventory but are excluded before collection.
\begin{table}[h]
\caption{Per-configuration main requests.}\label{tab:variant-inventory}\centering\small
\begin{tabular}{lrr}\toprule
Variant & Intended & Eligible\\\midrule
Canonical & 1,000 & 1,000\\ Event & 1,000 & 1,000\\ Complement $\neg A$ & 1,000 & 1,000\\ Query of event $B$ & 1,000 & 1,000\\ Intersection & 1,000 & 1,000\\ Conditional & 1,000 & 913\\ Binary Choice & 1,000 & 1,000\\ Permutation & 1,000 & 1,000\\ Rendering & 1,000 & 1,000\\ Opaque keys & 1,000 & 1,000\\ Irrelevant metadata & 1,000 & 1,000\\ Exact repeat & 1,000 & 1,000\\\midrule
Total & 12,000 & 11,913\\\bottomrule
\end{tabular}
\end{table}

\section{Additional Accuracy and Accounting}\label{app:fulltables}

\begin{table*}[t]
\caption{Canonical posterior accuracy: eight-stratum means with 95\% paired root-bootstrap intervals. Scores use valid canonical roots; all 1,000 intended roots enter decision loss. Full response counts and realized Brier scores appear in Appendix~\ref{app:fulltables}.}
\label{tab:accuracy}\centering\small
\begin{tabular}{lrrr}
\toprule
Configuration & Valid canonical roots & Posterior TV & Expected Brier\\
\midrule
Kev & 1,000 & \shortstack{0.2350\\{}[0.2278, 0.2423]} & \shortstack{0.6333\\{}[0.6229, 0.6436]}\\[5pt]
Base & 1,000 & \shortstack{0.3740\\{}[0.3655, 0.3824]} & \shortstack{0.7474\\{}[0.7379, 0.7571]}\\[5pt]
Posttrained & 1,000 & \shortstack{0.3202\\{}[0.3124, 0.3280]} & \shortstack{0.6938\\{}[0.6831, 0.7048]}\\[5pt]
Jev-1.13.0 & 990 & \shortstack{0.2677\\{}[0.2599, 0.2754]} & \shortstack{0.6560\\{}[0.6415, 0.6705]}\\
\bottomrule
\end{tabular}
\end{table*}
\subsection{Full paired comparisons and calibration}
Kev has the lowest aggregate canonical TV: 0.2350, compared with 0.3740 for base, 0.3202 for posttrained, and 0.2677 for Jev (Appendix Table~\ref{tab:accuracy}). The first three estimates use 1,000 valid roots; Jev uses 990. Paired differences favor Kev over base by $-0.1390$ [$-0.1494$, $-0.1286$] and over posttrained by $-0.0853$ [$-0.0942$, $-0.0761$]. Posttrained minus base is $-0.0537$ [$-0.0608$, $-0.0465$].

On 990 paired-valid roots, Jev minus Kev is 0.0335 [0.0222, 0.0448]. Jev minus base is $-0.1065$ [$-0.1190$, $-0.0940$], and Jev minus posttrained is $-0.0524$ [$-0.0633$, $-0.0413$]. These paired differences need not equal differences of the full marginal means because validity changes the comparison population.

Kev and Jev each have lower canonical TV in four of the eight strata (Appendix Table~\ref{tab:strata}); one Jev advantage is near a tie, 0.1858 versus 0.1866. Kev's aggregate advantage is concentrated in joint-attributes and noisy-sensor strata. On missing-rule joined tables, its TV is 0.4545 versus Jev's 0.1931. The post-hoc paired-valid analysis also splits 4--4 (Appendix~\ref{app:accelerated-offline}). Thus the aggregate advantage is not uniform dominance.

Expected log loss is finite for all original local canonical outputs. Jev has 20 infinite expected and 5 infinite realized log losses among its 990 valid roots. The post-hoc 15-bin exact-posterior calibration gaps are 0.0618, 0.1441, 0.0877, and 0.1952 for Kev, base, posttrained, and Jev. They compare confidence with the posterior mass of the selected class, avoiding outcome-draw noise without identifying per-root accuracy. Appendix~\ref{app:symmetric} retains the distinct prespecified sampled-outcome summaries.

Each configuration has 12,000 intended slots and 11,913 eligible requests; 87 conditionals are undefined. All three local configurations returned valid vectors on every eligible request. Jev has 11,849 valid, 62 rejected, and 2 unavailable responses. Table~\ref{tab:realized} gives the corresponding realized canonical Brier scores.
\begin{table}[h]
\caption{Realized canonical Brier: mean [95\% interval].}\label{tab:realized}\centering\small
\begin{tabular}{lr}\toprule
Configuration & Realized Brier\\\midrule
Kev & 0.6359 [0.6160, 0.6571]\\
Base & 0.7530 [0.7332, 0.7739]\\
Posttrained & 0.6918 [0.6696, 0.7137]\\
Jev-1.13.0 & 0.6469 [0.6094, 0.6851]\\\bottomrule
\end{tabular}\end{table}

\section{Collection Accounting}\label{app:collection}
Every main Jev request has one recorded transport attempt: 11,849 valid, 62 rejected after HTTP 200, and two unavailable after HTTP 403 and HTTP 520 respectively. These two statuses were ineligible for main retries. The protocol allowed at most six identical-payload attempts for HTTP 429, HTTP 529, or pre-response network failures, but no main request used that allowance. Two later one-shot service-restoration probes returned HTTP 200; their outputs are excluded and the original unavailable outcomes remain unchanged.

Main results required complete request accounting, an independent audit, version and exact-payload verification, relation-specific valid denominators, and consistency between the machine-readable results and manuscript. The 477-request development cohort is excluded from all estimates. The compact submission supplement contains selected scores, analysis products, code, and audit records. Complete raw-response archives are retained separately; the compact artifact alone does not replay the full collection from raw bytes.

\section{Decimal Boundary Handling}\label{app:numerics}
Pass classification uses the decimal values serialized in the responses, with inclusive comparisons to the specified tolerance. Binary floating-point subtraction can represent an exact decimal difference of 0.02 as slightly larger than 0.02; treating that as a failure would alter the intended rule. The corrected calculation and an independent implementation agree that 745 of 1,000 Jev exact-repeat pairs pass. This numerical correction changes neither responses nor tolerances. Earlier analyses are retained for traceability. All current counts use the decimal-domain comparison, and the joint criterion remains unmet. The extension's strict JSON parser likewise uses an inclusive decimal simplex tolerance of $10^{-6}$. One accepted vector $(0.333333,0.333333,0.333333)$ lies exactly on this boundary. The offline validator sums the stored decimal values exactly so binary roundoff does not reject it; probabilities and collection statuses remain unchanged. Canonical expected-cost calculations use the supplied vector without normalization. Relation-specific binary controllers use the reported scalar $p(A)$ and its logical complement, distinct from scoring the full canonical vector.

\section{Collection Continuity Diagnostic}\label{app:continuity}
Immutable receipts partitioned the main run into 174 pre-HTTP403 requests (172/1/1 valid/rejected/unavailable), 6,495 post-HTTP403 and pre-HTTP520 requests (6,452/42/1), and 5,244 post-HTTP520 requests (5,225/19/0). No admitted request was removed. Fixed-stratum, fixed-primitive differences were mixed around the 403. After the 520, the two supported Choice cells had higher TV, Brier, and regret, while Noul changes were small and mixed. The request order changed roots and generator support across windows, so this is a descriptive continuity check rather than a causal temporal comparison.

\section{An Action-Changing World}\label{app:example}
For a post-hoc illustration, we select a root by a fixed hash ordering among pairs where Jev's event answer strictly defers and its Choice answer strictly acts. This illustrates a mechanism, not its prevalence. The selected noisy-sensor world has prior counts $(67,85,9,79)$ and sensor likelihoods $(1/2,9/10,1,1/2)$. The cells' unnormalized masses are $(33.5,76.5,9,39.5)$, totaling $158.5$. The first two cells form event $A$, so $q(A)=110/158.5=220/317\approx0.694$. Jev returns $p(A)=0.84$ through Noul and $0.97$ through Choice, with the same state, instructions, and event definitions. At $c=0.10$, the $0.90$ action threshold separates these answers: Noul leads to deferral with loss 0.10, whereas Choice leads to answering yes with expected loss $97/317\approx0.306$. A probability shift of 0.13 thus increases loss by about 0.206 in this example. The supplement identifies the retained numerical reports and exact posterior for this example; its selection rule is stated above. The original request/response bytes and selection script are not included.

\paragraph{A complete stored request.}
The following is a separate noisy-sensor Event request from the headroom cohort. It shows every payload field, with line wrapping for readability; it is not a reconstruction of the introductory request. The question defines event $A$ as categories \texttt{type\_0} and \texttt{type\_1}. Attributes, counts, and evidence likelihoods are supplied in three tables joined on \texttt{cell}. The exact payload and its binding to the stored request appear in \path{docs/request_example/} in the supplement.
\begingroup
\small
\begin{verbatim}
model: jev-latest
questions.q.type: noul
questions.q.instructions:
Estimate the probability that category is one of ['type_0', 'type_1'].
state:
Choose a type proportional to prior counts; observe sensor reading R. The likelihood column is
P(R|type). All unlisted evidence is absent. Weights need not sum to one. Join the three tables on
cell; sample cell mass is prior_count times evidence_likelihood.
cell,a,b,category
0,False,False,type_0
1,False,True,type_1
2,True,False,type_2
3,True,True,type_3
---
cell,prior_count
0,2
1,19
2,16
3,10
---
cell,evidence_likelihood
0,1/5
1,0
2,3/5
3,2/5
\end{verbatim}
\endgroup

The likelihood readout uses the same kind of state and event question, with explicit answer IDs assigned to options. It reads the model's next-token probabilities for those IDs and normalizes within that set; this differs from asking the model to generate numerical percentages.

\section{Decision Geometry}\label{app:geometry}
For 0--1 class loss and defer cost $c$, the Bayes risk and headroom below always defer are
\begin{align*}
L_c^*(q)&=\min\{c,1-\lVert q\rVert_\infty\},\\
H_c(q)&=\max\{0,\lVert q\rVert_\infty-(1-c)\}.
\end{align*}
For a normalized vector, the controller acts exactly when $\max_i p_i\ge 1-c$. For posterior TV $\delta=\operatorname{TV}(p,q)$, its regret to the oracle is at most $2\delta$; an act/defer disagreement sharpens this to $\delta$. If $|p_{(1)}-(1-c)|>\operatorname{TV}(p,p')$, two representations have the same act/defer status, and if $p_{(1)}-p_{(2)}>2\operatorname{TV}(p,p')$, they have the same maximizing class. Finally, aligning separate reports $x=\widehat P(A)$ and $z=\widehat P(A^c)$ gives
\[
\operatorname{TV}((1-x,x),(z,1-z))=|x+z-1|,
\]
so the complement residual lower-bounds the worse representation-specific posterior error by half its magnitude. These bounds use TV to evaluator truth when claiming regret; pairwise representation TV alone does not bound true downstream loss. To see the regret bound, let $a_p,a_q$ minimize predicted and true expected loss. Insert and subtract their predicted losses. Predicted optimality makes the intervening difference nonpositive, while each class-action loss changes by at most $\delta$ and defer loss does not change. This gives $2\delta$ in general and $\delta$ if one selected action is defer. The margin claims follow from $|p_i-p'_i|\leq\operatorname{TV}(p,p')$ for every coordinate. The complement bound is the triangle inequality for the two aligned binary distributions and their common truth.

\section{Symmetric Post-Hoc Checks}\label{app:symmetric}
For reference, the prespecified 15-bin sampled-outcome ECE values are 0.0691, 0.1691, 0.1176, and 0.1778 for Kev, base, posttrained, and Jev. These use one realized outcome per root and are distinct from the post-hoc exact-posterior summaries below.
Applying the same specified validity threshold, relation tolerances, and lower-bound rule to all four complete configurations found that none met the broad conjunction (Table~\ref{tab:symmetric}). This descriptive extension is useful context for the prospectively specified Jev evaluation: the benchmark exposes different relation failures across every configuration rather than a defect unique to one interface. The exact-posterior calibration column bins the reported top-class probability against the mean exact posterior mass of that selected class. It is distinct from sampled-outcome ECE and was not a prespecified primary endpoint.

\begin{table}[h]
\caption{Symmetric all-configuration check. Exact-$q$ gap is a 15-bin top-class calibration summary against evaluator posterior mass.}
\label{tab:symmetric}\centering\small
\resizebox{\columnwidth}{!}{\begin{tabular}{lrrc}
\toprule
Configuration & Native validity & Exact-$q$ gap & Operational criterion\\
\midrule
Kev & 1.0000 & 0.0618 & No\\
Base & 1.0000 & 0.1441 & No\\
Posttrained & 1.0000 & 0.0877 & No\\
Jev-1.13.0 & 0.9946 & 0.1952 & No\\
\bottomrule
\end{tabular}}
\end{table}

\section{Full Relation Estimates}\label{app:allrelations}
Tables~\ref{tab:relations-kev}, \ref{tab:relations-base}, \ref{tab:relations-post}, and \ref{tab:relations-jev} report all ten relations, including those not displayed in the main comparison. Values summarize the existing primary relation records; this consolidated presentation adds no model calls. Product eligibility excludes zero conditioning mass. Missing or rejected constituent responses reduce valid support further. Intervals use the specified 5,000 root bootstrap samples and are pointwise. The supplied CSV contains the unrounded values and full denominator fields.
\begin{table*}[t]
\caption{All continuous relation residuals for Kev. Estimates and intervals preserve each relation's paired-valid support.}
\label{tab:relations-kev}\centering\small
\begin{tabular}{llrr}\toprule
Relation & Mean [95\% interval] & Valid roots & Eligible roots\\\midrule
Complement & 0.5055 [0.4971, 0.5138] & 1000 & 1000\\
Event/Choice interface & 0.0629 [0.0610, 0.0648] & 1000 & 1000\\
Coarsening & 0.3554 [0.3489, 0.3620] & 1000 & 1000\\
Intersection bounds & 0.1553 [0.1507, 0.1598] & 1000 & 1000\\
Product & 0.1328 [0.1300, 0.1356] & 913 & 913\\
Exact repeat & 0.0000 [0.0000, 0.0000] & 1000 & 1000\\
Option permutation TV & 0.1143 [0.1104, 0.1181] & 1000 & 1000\\
Rendering TV & 0.0891 [0.0859, 0.0923] & 1000 & 1000\\
Opaque-key TV & 0.0797 [0.0779, 0.0815] & 1000 & 1000\\
Irrelevant-metadata TV & 0.0636 [0.0614, 0.0660] & 1000 & 1000\\
\bottomrule\end{tabular}\end{table*}

\begin{table*}[t]
\caption{All continuous relation residuals for Base. Estimates and intervals preserve each relation's paired-valid support.}
\label{tab:relations-base}\centering\small
\begin{tabular}{llrr}\toprule
Relation & Mean [95\% interval] & Valid roots & Eligible roots\\\midrule
Complement & 0.1155 [0.1133, 0.1177] & 1000 & 1000\\
Event/Choice interface & 0.2138 [0.2111, 0.2165] & 1000 & 1000\\
Coarsening & 0.2451 [0.2420, 0.2481] & 1000 & 1000\\
Intersection bounds & 0.0814 [0.0807, 0.0821] & 1000 & 1000\\
Product & 0.2682 [0.2673, 0.2691] & 913 & 913\\
Exact repeat & 0.0000 [0.0000, 0.0000] & 1000 & 1000\\
Option permutation TV & 0.3224 [0.3150, 0.3294] & 1000 & 1000\\
Rendering TV & 0.1329 [0.1315, 0.1343] & 1000 & 1000\\
Opaque-key TV & 0.0541 [0.0532, 0.0551] & 1000 & 1000\\
Irrelevant-metadata TV & 0.0261 [0.0255, 0.0268] & 1000 & 1000\\
\bottomrule\end{tabular}\end{table*}

\begin{table*}[t]
\caption{All continuous relation residuals for Posttrained. Estimates and intervals preserve each relation's paired-valid support.}
\label{tab:relations-post}\centering\small
\begin{tabular}{llrr}\toprule
Relation & Mean [95\% interval] & Valid roots & Eligible roots\\\midrule
Complement & 0.8000 [0.7982, 0.8017] & 1000 & 1000\\
Event/Choice interface & 0.1126 [0.1085, 0.1168] & 1000 & 1000\\
Coarsening & 0.2836 [0.2776, 0.2896] & 1000 & 1000\\
Intersection bounds & 0.0358 [0.0342, 0.0375] & 1000 & 1000\\
Product & 0.1406 [0.1389, 0.1424] & 913 & 913\\
Exact repeat & 0.0000 [0.0000, 0.0000] & 1000 & 1000\\
Option permutation TV & 0.3470 [0.3376, 0.3562] & 1000 & 1000\\
Rendering TV & 0.1026 [0.1002, 0.1050] & 1000 & 1000\\
Opaque-key TV & 0.2105 [0.2074, 0.2136] & 1000 & 1000\\
Irrelevant-metadata TV & 0.0767 [0.0750, 0.0783] & 1000 & 1000\\
\bottomrule\end{tabular}\end{table*}

\begin{table*}[t]
\caption{All continuous relation residuals for Jev-1.13.0. Estimates and intervals preserve each relation's paired-valid support.}
\label{tab:relations-jev}\centering\small
\begin{tabular}{llrr}\toprule
Relation & Mean [95\% interval] & Valid roots & Eligible roots\\\midrule
Complement & 0.0929 [0.0890, 0.0967] & 1000 & 1000\\
Event/Choice interface & 0.1938 [0.1873, 0.2002] & 1000 & 1000\\
Coarsening & 0.1751 [0.1690, 0.1813] & 990 & 1000\\
Intersection bounds & 0.0200 [0.0173, 0.0227] & 999 & 1000\\
Product & 0.0836 [0.0804, 0.0868] & 912 & 913\\
Exact repeat & 0.0188 [0.0176, 0.0200] & 1000 & 1000\\
Option permutation TV & 0.0534 [0.0503, 0.0567] & 976 & 1000\\
Rendering TV & 0.0563 [0.0534, 0.0592] & 977 & 1000\\
Opaque-key TV & 0.0644 [0.0614, 0.0674] & 976 & 1000\\
Irrelevant-metadata TV & 0.0469 [0.0444, 0.0493] & 979 & 1000\\
\bottomrule\end{tabular}\end{table*}

\section{Accounting Views and Stratum Estimates}
Table~\ref{tab:populations} distinguishes the four accounting views; Table~\ref{tab:strata} gives canonical posterior TV for each generator--encoding stratum.
\begin{table}[t]
\caption{Four accounting views retained for every configuration.}
\label{tab:populations}\centering\small
\begin{tabular}{p{0.25\columnwidth}p{0.64\columnwidth}}
\toprule
View & Question\\\midrule
Eligible & Was the semantic transformation defined?\\
Native valid & Did the returned object satisfy its interface contract?\\
Paired valid & Are all vectors needed for this contrast scoreable?\\
Intended pipeline & What loss results when failure triggers defer?\\
\bottomrule
\end{tabular}
\end{table}
\begin{table*}[t]
\caption{Canonical posterior TV by prespecified generator--encoding stratum. Every specified stratum is shown, including reversals of aggregate ordering.}
\label{tab:strata}\centering\small
\resizebox{\textwidth}{!}{\begin{tabular}{lrrrr}
\toprule
Stratum & Kev & Base & Posttrained & Jev\\
\midrule
Finite population / joined tables & 0.2696 & 0.4061 & 0.2381 & 0.2429\\
Finite population / nested records & 0.1866 & 0.3905 & 0.3408 & 0.1858\\
Joint attributes / joined tables & 0.1764 & 0.3097 & 0.2468 & 0.4017\\
Joint attributes / nested records & 0.1868 & 0.2937 & 0.3489 & 0.3831\\
Missing rule / joined tables & 0.4545 & 0.4111 & 0.3430 & 0.1931\\
Missing rule / nested records & 0.1598 & 0.3614 & 0.3516 & 0.1272\\
Noisy sensor / joined tables & 0.2002 & 0.4230 & 0.2871 & 0.2146\\
Noisy sensor / nested records & 0.2457 & 0.3961 & 0.4054 & 0.3935\\
\midrule
Eight-stratum macro & 0.2350 & 0.3740 & 0.3202 & 0.2677\\
\bottomrule
\end{tabular}}
\end{table*}
\section{Service and Channel Diagnostics}\label{app:diagnostics}\label{app:channels}
Jev returned 11,849 valid responses, 62 rejected responses, and 2 unavailable outcomes across its 11,913 eligible main requests. All returned responses identified Jev-1.13.0. Native validity is 0.9946 and service availability is 0.9998. Rejected and unavailable requests are excluded from probability scores but take the defer action in loss on all intended roots. This accounting keeps failures in the evaluated pipeline. Because low-cost deferral can lower loss by reducing coverage, loss must be interpreted together with validity and deferral.

A post-hoc serialization sensitivity normalizes 57 rejected Choice vectors with finite, nonnegative entries and total-mass error no greater than 0.0100. Counting these as parseable would raise effective validity to 11,906/11,913 (0.9994), consistent with rounding near the simplex boundary. Primary responses remain rejected and unscored, and this sensitivity does not change failure of the joint benchmark criterion. Appendix~\ref{app:numerics} documents inclusive decimal-threshold handling so that a numerical boundary is not mistaken for a scientific effect.

The likelihood proxies have different normalization behavior. Mean mass on the allowed answer tokens is 0.0110 for base (median 0.0099) and 0.6570 for posttrained (median 0.6932). A sharp distribution conditional on a selected answer set can thus discard most of the model's full-vocabulary probability. Such a proxy should not be described as an unconditional semantic-event posterior.

A separate diagnostic asked posttrained to generate JSON probabilities. All 400 outputs satisfied the strict parser without repair. Relative to its likelihood channel, the probability gap was 0.1737 [0.1645, 0.1834] and the action-change rate was 0.5796 [0.5445, 0.6146]. This comparison tests two elicitation channels rather than a native probability head. Another diagnostic removed context: Kev and posttrained changed probabilities by 0.2964 [0.2875, 0.3052] and 0.2269 [0.2230, 0.2308] respectively on 200 roots and 400 paired cases each. Removing information is not an equivalence test, so those responses do not enter coherence claims.

\paragraph{Signed complement mass (post hoc).}
Recomputing $s=p(A)+p(\neg A)$ from the preserved fraction exports gives Table~\ref{tab:signed-complement}. Every Kev and posttrained pair has excess mass; the posttrained median is 1.8151, with range 1.5743--1.9150. These summaries distinguish systematic excess from cancellation of high and low sums. This historical diagnostic alone does not isolate answer-label preference, event interpretation, arithmetic, or readout construction. The separate joint surface-order and answer-ID intervention in Appendix~\ref{app:prospective-stress} measures sensitivity to that recoding without identifying a unique internal mechanism. No contextual recalibration was performed.
\begin{table}[ht]
\centering\small
\setlength{\tabcolsep}{4pt}
\caption{Signed complement sums on 1,000 pairs per configuration. Post-hoc summaries; all configurations retained.}\label{tab:signed-complement}
\begin{tabular}{lrrrr}
\toprule
Configuration & Mean & Minimum & Maximum & $s>1$\\
\midrule
Kev & 1.5055 & 1.0485 & 1.8153 & 1000\\
Base & 1.1103 & 0.9344 & 1.3733 & 861\\
Posttrained & 1.8000 & 1.5743 & 1.9150 & 1000\\
Jev & 1.0641 & 0.6700 & 1.3900 & 786\\
\bottomrule
\end{tabular}
\end{table}

\section{Cross-Family Extension Details}\label{app:extension}
\subsection{Readout comparison on the shared subset}
Table~\ref{tab:extension} evaluates the four new readouts on a shared subset. Both likelihood readouts return valid vectors throughout, yet complement gaps remain large: 0.6469 for Mistral and 0.2048 for OLMo. At $c=0.10$, Mistral's canonical loss is 0.1734; OLMo likelihood always defers. This extends the observation that validity alone does not ensure coherent reports or beneficial decisions.

Generated JSON has a different limitation: only 59 Mistral and 60 OLMo canonical outputs are valid. OLMo's canonical mean excludes one empty stratum, and its complement mean covers five of eight strata. These are conditional summaries, not full-population accuracy estimates. On 59 paired-valid roots, Mistral likelihood-minus-JSON TV is 0.1145 [0.0454, 0.1855]; OLMo's contrast is $-0.0164$ [$-0.0902$, $0.0558$] on 60 roots. Mistral JSON's lower pipeline loss also accompanies greater deferral, including invalid outputs.

Event and Choice compile to identical model-facing inputs for these four readouts. Their zero gap is an identity control, not an independent test of Jev's native interface contrast. The tables below report all six contrasts, ten relations, costs, and support cells; the JSON--JSON canonical comparison has only 21 roots across four strata. The extension broadens the tested configurations, but does not add independent worlds or schema-constrained generation.

\begin{table*}[t]
\caption{Four new configurations on the same 200-world subset. Validity uses all eligible requests; parentheses give valid canonical or paired-valid roots. Loss and defer rate retain all 200 roots at $c=0.10$. Rejected outputs defer. Valid-only means average nonempty strata; full support and pointwise intervals are in Appendix~\ref{app:extension}.}
\label{tab:extension}\centering\small
\begin{tabular}{lrrrrr}
\toprule
Configuration & Valid requests & Canonical TV ($n$) & Complement gap ($n$) & Loss & Defer\\
\midrule
Mistral likelihood & 2385/2385 & 0.4799 (200) & 0.6469 (200) & 0.1734 & 0.8900\\
Mistral JSON & 1454/2385 & 0.3937 (59) & 0.4848 (139) & 0.1119 & 0.9650\\
OLMo likelihood & 2385/2385 & 0.4261 (200) & 0.2048 (200) & 0.1000 & 1.0000\\
OLMo JSON & 1248/2385 & 0.4220 (60) & 0.2014 (103) & 0.1142 & 0.9700\\
\bottomrule
\end{tabular}
\end{table*}

The extension evaluates two instruction-tuned checkpoints through two readouts each. Mistral-7B-Instruct-v0.3 uses revision \texttt{c170c708c41dac9275d15a8fff4eca08d52bab71}; OLMo-2-1124-7B-Instruct uses \texttt{470b1fba1ae01581f270116362ee4aa1b97f4c84}. Exact tokenizer and asset inventories, runtime receipts, and request records are retained in separate verified evidence backups. Collection used one NVIDIA A100-SXM4-80GB, Python 3.13.5, PyTorch 2.8.0 with CUDA 12.8, and Transformers 5.17.0. Both models ran in BF16 without quantization, CPU offload, or fitted calibration.

The main subset contains 200 historical roots, 25 in each generator--encoding stratum. Each readout has 2,400 intended request slots, of which 15 conditionals are not applicable; 2,385 requests remain. A separate 477-case development allocation per readout checked tokenization, runtime and timing and contributes no evaluation result. Two models, two readouts, and both phases produced 11,448 terminal attempts, with zero retries. The likelihood readout normalizes only the admitted single-token answer IDs and retains the total vocabulary mass on those IDs. The generated readout uses greedy decoding with at most 128 new tokens and a strict JSON parser. Parse and simplex failures are rejected without extracting, normalizing, or repairing a vector.

\begin{table}[t]
\caption{Strict generated-JSON rejection reasons across all 2,385 main requests per model. Categories are mutually exclusive parser outcomes; none is repaired. These are protocol-validity failures, distinct from relation residuals on valid reports.}
\label{tab:extension-rejections}\centering\small
\begin{tabular}{lrr}
\toprule
Reason & Mistral & OLMo\\
\midrule
Invalid JSON & 528 & 4\\
Wrong key support & 0 & 536\\
Nonnumeric probability & 21 & 0\\
Trailing content & 8 & 0\\
Out-of-range probability & 62 & 1\\
Simplex sum mismatch & 312 & 596\\
\midrule
Total rejected & 931 & 1137\\
\bottomrule
\end{tabular}
\end{table}

For every new model/readout, the Event request for event $A$, binary Choice and exact repeat have byte-identical compiled prompts, token IDs, answer mappings and generation limits on all 200 roots. Their paired-valid vectors match exactly (200 pairs for each likelihood readout, 140 for Mistral JSON, 196 for OLMo JSON). Their zero residuals are therefore deterministic identity controls. The extension does not replicate Jev's native Noul-versus-Choice intervention; complement, coarsening and the other nonidentical transformations remain distinct tests.

Table~\ref{tab:extension-rejections} separates parser outcomes. Sixteen rejected Mistral responses generated exactly 128 tokens; no OLMo response reached that count. Median generated lengths were 25 and 21 tokens, respectively. These descriptive diagnostics do not assign a causal mechanism to a rejection. Strict JSON reporting here is an unconstrained generated-text channel, not a provider-enforced typed schema.

Every main collection completed its intended allocation. Development plus main collection occupied 44.77 minutes for Mistral and 41.54 minutes for OLMo, including model loading; acquisition time is separate. Available runtime summaries accompany the released exports; full collection source and original development/main archives are not included.

The extension's six unordered configuration contrasts were specified before these outputs. Probability-score contrasts use roots valid for both configurations; decision contrasts include all 200 roots, assigning rejected canonical outputs to defer. Marginal valid-only estimates average the nonempty strata equally. If a stratum has no valid outputs, it cannot contribute to that probability estimate; it remains present in decisions on all intended roots. The full support inventory accompanies each relation and contrast. Intervals are pointwise paired root-bootstrap summaries, not multiplicity-adjusted evidence of superiority. Historical subset results are contextual: prompts, precision, channel and collection time differ, so they do not identify an architectural effect or constitute a new-data replication.

Tables~\ref{tab:extension-canonical}--\ref{tab:extension-support} report the complete subset estimates, intervals and selected support counts. The machine-readable independent support audit retains all relation-specific support, including zero-valid cells. OLMo JSON has no valid canonical output in the missing-rule joined-table stratum. Its complement estimate spans five strata, and its intersection and product estimates span only three. The paired JSON--JSON canonical comparison uses 21 roots across four strata. These valid-only quantities describe different conditional populations; they must not be read as eight-stratum full-pipeline rankings.

\begin{table*}[t]
\caption{Canonical subset results with 95\% pointwise root-bootstrap intervals. Probability scores use the stated valid support; expected decision loss uses all 200 roots at $c=0.10$. Historical rows reuse earlier responses on exactly this subset and are not a controlled architecture comparison.}
\label{tab:extension-canonical}\centering\small
\begin{tabular}{lrrrr}
\toprule
Configuration & $n$ valid & TV & Expected Brier & Expected loss\\
\midrule
Mistral likelihood & 200 & 0.4799 [0.4517, 0.5099] & 0.9696 [0.9275, 1.0132] & 0.1734 [0.1531, 0.1942]\\
Mistral JSON & 59 & 0.3937 [0.3506, 0.4358] & 0.8126 [0.7548, 0.8725] & 0.1119 [0.1001, 0.1271]\\
OLMo likelihood & 200 & 0.4261 [0.3955, 0.4583] & 0.8566 [0.8224, 0.8913] & 0.1000 [0.1000, 0.1000]\\
OLMo JSON & 60 & 0.4220 [0.3767, 0.4666] & 0.8224 [0.7598, 0.8884] & 0.1142 [0.1020, 0.1299]\\
Kev (historical) & 200 & 0.2349 [0.2179, 0.2522] & 0.6239 [0.6028, 0.6445] & 0.1079 [0.1016, 0.1149]\\
Base (historical) & 200 & 0.3785 [0.3601, 0.3967] & 0.7474 [0.7250, 0.7698] & 0.1000 [0.1000, 0.1000]\\
Posttrained (historical) & 200 & 0.3296 [0.3111, 0.3477] & 0.7011 [0.6763, 0.7257] & 0.1000 [0.1000, 0.1000]\\
Jev (historical) & 200 & 0.2645 [0.2475, 0.2818] & 0.6448 [0.6123, 0.6761] & 0.1422 [0.1240, 0.1603]\\
\bottomrule
\end{tabular}
\end{table*}

\begin{table*}[t]
\caption{Mistral relation residuals with pointwise intervals. Each readout has 200 intended roots, except product (185 eligible). $n$ counts complete valid constituents. Missing strata are excluded only from valid-only aggregates; the machine-readable support inventory identifies every exclusion.}
\label{tab:extension-relations-mistral7b-instruct}\centering\small
\begin{tabular}{lrrrr}
\toprule
Relation & $n$ likelihood & Likelihood residual & $n$ JSON & JSON residual\\
\midrule
Event/Choice & 200 & 0.0000 [0.0000, 0.0000] & 140 & 0.0000 [0.0000, 0.0000]\\
Complement & 200 & 0.6469 [0.6340, 0.6600] & 139 & 0.4848 [0.4423, 0.5269]\\
Coarsening & 200 & 0.2882 [0.2730, 0.3034] & 51 & 0.2055 [0.1664, 0.2448]\\
Intersection bound & 200 & 0.1091 [0.1052, 0.1131] & 124 & 0.2184 [0.1904, 0.2490]\\
Product & 185 & 0.1575 [0.1527, 0.1625] & 146 & 0.3279 [0.2957, 0.3626]\\
Permutation & 200 & 0.6167 [0.5891, 0.6432] & 39 & 0.4039 [0.3559, 0.4553]\\
Rendering & 200 & 0.1363 [0.1291, 0.1441] & 31 & 0.1197 [0.0719, 0.1636]\\
Opaque keys & 200 & 0.1347 [0.1276, 0.1420] & 44 & 0.0772 [0.0414, 0.1209]\\
Irrelevant metadata & 200 & 0.1398 [0.1324, 0.1473] & 47 & 0.0895 [0.0486, 0.1329]\\
Exact repeat & 200 & 0.0000 [0.0000, 0.0000] & 140 & 0.0000 [0.0000, 0.0000]\\
\bottomrule
\end{tabular}
\end{table*}

\begin{table*}[t]
\caption{OLMo relation residuals with pointwise intervals. Each readout has 200 intended roots, except product (185 eligible). $n$ counts complete valid constituents. Missing strata are excluded only from valid-only aggregates; the machine-readable support inventory identifies every exclusion.}
\label{tab:extension-relations-olmo2-7b-instruct}\centering\small
\begin{tabular}{lrrrr}
\toprule
Relation & $n$ likelihood & Likelihood residual & $n$ JSON & JSON residual\\
\midrule
Event/Choice & 200 & 0.0000 [0.0000, 0.0000] & 196 & 0.0000 [0.0000, 0.0000]\\
Complement & 200 & 0.2048 [0.2012, 0.2084] & 103 & 0.2014 [0.1742, 0.2297]\\
Coarsening & 200 & 0.4655 [0.4599, 0.4711] & 58 & 0.2456 [0.1977, 0.2937]\\
Intersection bound & 200 & 0.0062 [0.0053, 0.0071] & 46 & 0.2206 [0.1730, 0.2713]\\
Product & 185 & 0.0984 [0.0958, 0.1010] & 45 & 0.3613 [0.3201, 0.4026]\\
Permutation & 200 & 0.4890 [0.4633, 0.5152] & 34 & 0.4036 [0.3343, 0.4757]\\
Rendering & 200 & 0.1278 [0.1231, 0.1328] & 21 & 0.2084 [0.1540, 0.2684]\\
Opaque keys & 200 & 0.1741 [0.1703, 0.1779] & 40 & 0.0641 [0.0304, 0.1011]\\
Irrelevant metadata & 200 & 0.0651 [0.0617, 0.0687] & 44 & 0.0694 [0.0360, 0.1046]\\
Exact repeat & 200 & 0.0000 [0.0000, 0.0000] & 196 & 0.0000 [0.0000, 0.0000]\\
\bottomrule
\end{tabular}
\end{table*}

\begin{table*}[t]
\caption{Expected canonical decision loss for the complete fixed cost grid, all 200 roots per configuration. Rejected and unavailable outputs defer. Always-defer loss equals the column cost.}
\label{tab:extension-costs}\centering\small
\begin{tabular}{lrrrrr}
\toprule
Configuration & $c=.02$ & $c=.05$ & $c=.10$ & $c=.20$ & $c=.40$\\
\midrule
Mistral likelihood & 0.0200 & 0.0581 & 0.1734 & 0.3124 & 0.5635\\
Mistral JSON & 0.0200 & 0.0523 & 0.1119 & 0.2140 & 0.4304\\
OLMo likelihood & 0.0200 & 0.0500 & 0.1000 & 0.2000 & 0.4304\\
OLMo JSON & 0.0299 & 0.0640 & 0.1142 & 0.2177 & 0.4195\\
Kev (historical) & 0.0200 & 0.0500 & 0.1079 & 0.1999 & 0.3913\\
Base (historical) & 0.0200 & 0.0500 & 0.1000 & 0.2000 & 0.3844\\
Posttrained (historical) & 0.0200 & 0.0500 & 0.1000 & 0.2000 & 0.4012\\
Jev (historical) & 0.0212 & 0.0784 & 0.1422 & 0.2568 & 0.4116\\
\bottomrule
\end{tabular}
\end{table*}

\begin{table*}[t]
\caption{All six planned configuration contrasts (left minus right). TV uses paired-valid canonical roots; loss uses all 200 roots at $c=0.10$. Intervals are pointwise, without multiplicity-adjusted superiority claims. The full result file also includes Brier and the other fixed costs.}
\label{tab:extension-contrasts}\centering\small
\begin{tabular}{lrrr}
\toprule
Contrast & $n$ TV & TV difference & Expected-loss difference\\
\midrule
Mistral likelihood -- Mistral JSON & 59 & 0.1145 [0.0454, 0.1855] & 0.0615 [0.0342, 0.0871]\\
Mistral likelihood -- OLMo likelihood & 200 & 0.0538 [0.0331, 0.0750] & 0.0734 [0.0531, 0.0942]\\
Mistral likelihood -- OLMo JSON & 60 & 0.0790 [0.0015, 0.1582] & 0.0591 [0.0368, 0.0816]\\
Mistral JSON -- OLMo likelihood & 59 & -0.0780 [-0.1505, -0.0077] & 0.0119 [0.0001, 0.0271]\\
Mistral JSON -- OLMo JSON & 21 & -0.1047 [-0.1629, -0.0506] & -0.0024 [-0.0226, 0.0179]\\
OLMo likelihood -- OLMo JSON & 60 & -0.0164 [-0.0902, 0.0558] & -0.0142 [-0.0299, -0.0020]\\
\bottomrule
\end{tabular}
\end{table*}

\begin{table*}[t]
\caption{Valid-root support by the eight fixed strata (25 intended roots each): finite population (F), joint attributes (J), missing rule (M), and noisy sensor (N), each in joined-table (T) and nested-record (R) form. Zero-support cells are excluded from valid-only means; all cells remain in decision loss. Canonical and Event/Choice supports need not coincide.}
\label{tab:extension-support}\centering\small
\begin{tabular}{lrrrrrrrr}
\toprule
Configuration and estimand & F/T & F/R & J/T & J/R & M/T & M/R & N/T & N/R\\
\midrule
Mistral likelihood, canonical & 25 & 25 & 25 & 25 & 25 & 25 & 25 & 25\\
Mistral likelihood, Event/Choice & 25 & 25 & 25 & 25 & 25 & 25 & 25 & 25\\
Mistral JSON, canonical & 8 & 1 & 8 & 2 & 15 & 16 & 8 & 1\\
Mistral JSON, Event/Choice & 25 & 12 & 18 & 7 & 24 & 24 & 19 & 11\\
OLMo likelihood, canonical & 25 & 25 & 25 & 25 & 25 & 25 & 25 & 25\\
OLMo likelihood, Event/Choice & 25 & 25 & 25 & 25 & 25 & 25 & 25 & 25\\
OLMo JSON, canonical & 2 & 6 & 6 & 9 & 0 & 21 & 8 & 8\\
OLMo JSON, Event/Choice & 25 & 22 & 25 & 25 & 25 & 24 & 25 & 25\\
\bottomrule
\end{tabular}
\end{table*}

\section{Additional Related-Work Context}\label{app:related}
Readout and representation sensitivity are established: paraphrases, answer slots, context, and verbalized-versus-token readout can change confidence conclusions \citep{mahaut2024confidence,kim2026protocol}. Concurrent Jev studies cover task calibration, review routing, and factuality judging \citep{rafe2026jev,huang2026jev,ibrahim2026decision}, option-name/rubric reassignment \citep{sun2026typesafe}, request configuration and question/context ablations \citep{zhang2026different,guo2026askjev}, answer-preserving context additions \citep{xu2026jevout}, and semantic selections with downstream quantities \citep{deng2026scientific}. Belief--choice consistency and policy-dependent elicitation further motivate separating reported probabilities from decisions \citep{yamin2026beliefs,kudum2026policies}. To our knowledge, these studies do not jointly evaluate exact finite-world posteriors, validated event relations, cross-interface alignment, served failures, and paired decision regret. Our contribution is their combination in one benchmark, rather than any one test in isolation.

\section{Post-Hoc Error Bounds and Fixed Policies}\label{app:posthoc}
This analysis reuses preserved predictions; it does not modify the primary collection or repair rejected outputs. Its specification was recorded after descriptive disagreement-to-error ratios had been inspected. It retains every configuration, the five fixed costs, and 5,000 shared eight-stratum root-bootstrap replicates (seed 2026092408). All intervals are pointwise descriptive estimates. They quantify sampling over these worlds, not repeated API collection or generalization to model families. The evaluator reconstructs $q(A)$ from public counts and likelihoods and checks exact agreement with all aligned evaluator targets. All 1,000 Event/Choice, Event/repeat, and Event/complement pairs are complete for each configuration.

For a valid report $p$, the controller minimizes $(p,1-p,c)$ over (no, yes, defer), in that order at ties. The evaluator retains \emph{all} minimizers under exact rational $q$ when deciding strict suboptimality. Different actions can both be Bayes-optimal; an action flip alone supplies no strictly positive universal regret bound. Policies receive only probability reports and the declared cost. Evaluator truth appears only in scoring.

For Eq.~\ref{eq:absolute-certificate}, sorting the two reports gives two cases: if $q$ lies between them, the two distances sum to their separation; otherwise their sum is the separation plus twice the distance to the nearer endpoint. Expanding the squares around $m$ proves Eq.~\ref{eq:squared-certificate}. These checks concern average pair error, not canonical multiclass TV. Expected Bernoulli Brier adds $q(1-q)$; the two-coordinate categorical convention multiplies the complete identity by two.

For Jev, the absolute-error lower bound accounts for 0.522 of mean pair error; for Kev, 0.107. The corresponding fractions are 0.416 for base and 0.149 for posttrained. Squared-error fractions are 0.232, 0.014, 0.131, and 0.031 in the same order. Each fraction is the ratio of the mean lower bound to mean pair error. Kev's smaller interface gap therefore certifies a smaller share of its binary error, despite its lower canonical TV. 

\begin{table*}[t]
\caption{Post-hoc absolute-error decomposition for Event/Choice: mean and pointwise 95\% stratified root-bootstrap intervals. The disagreement lower bound plus error beyond the bound equals mean pair error before rounding.}
\label{tab:certificate-components}\centering\small
\begin{tabular}{lrrr}\toprule
Configuration & Pair absolute error & Disagreement lower bound & Error beyond bound\\\midrule
Kev & 0.29397 [0.28537, 0.30287] & 0.03143 [0.03048, 0.03240] & 0.26254 [0.25384, 0.27140]\\
Base & 0.25692 [0.24884, 0.26514] & 0.10689 [0.10553, 0.10816] & 0.15004 [0.14149, 0.15864]\\
Posttrained & 0.37758 [0.36562, 0.38971] & 0.05632 [0.05423, 0.05836] & 0.32126 [0.30941, 0.33318]\\
Jev & 0.18567 [0.18023, 0.19105] & 0.09688 [0.09362, 0.10010] & 0.08879 [0.08376, 0.09393]\\
\bottomrule\end{tabular}\end{table*}

\begin{table*}[t]
\caption{Post-hoc primary reporting contrasts at $c=0.10$: expected loss difference with pointwise 95\% intervals. Negative favors the first policy. Random interface selects Event or Choice with equal probability. No multiplicity-adjusted significance claim is made.}
\label{tab:policy-contrasts}\centering\small
\begin{tabular}{lrr}\toprule
Configuration & Mean minus Random interface & Agreement minus mean\\\midrule
Kev & -0.00959 [-0.01347, -0.00581] & -0.01095 [-0.01503, -0.00733]\\
Base & 0.00000 [0.00000, 0.00000] & 0.00000 [0.00000, 0.00000]\\
Posttrained & -0.05297 [-0.05964, -0.04634] & -0.01816 [-0.02399, -0.01291]\\
Jev & -0.03463 [-0.03886, -0.03049] & -0.00236 [-0.00411, -0.00077]\\
\bottomrule\end{tabular}\end{table*}

\begin{table*}[t]
\caption{Post-hoc binary expected loss across all fixed costs. Every configuration has 1,000 intended roots and complete inputs. Event/Choice use one query; mean/agreement and repeat controls use two. Always defer has loss $c$. Oracle is evaluator-only. Full pointwise intervals and coverage remain in the machine-readable results.}
\label{tab:binary-cost-grid}\centering\small
\begin{tabular}{llrrrrrrr}\toprule
Configuration & $c$ & Event & Choice & Mean & Agreement & Repeat mean & Repeat agreement & Oracle\\\midrule
Kev & 0.02 & 0.0200 & 0.0200 & 0.0200 & 0.0200 & 0.0200 & 0.0200 & 0.0165\\
Kev & 0.05 & 0.0512 & 0.0516 & 0.0506 & 0.0501 & 0.0512 & 0.0512 & 0.0411\\
Kev & 0.10 & 0.1669 & 0.1579 & 0.1528 & 0.1419 & 0.1669 & 0.1669 & 0.0813\\
Kev & 0.20 & 0.3128 & 0.3052 & 0.3122 & 0.2885 & 0.3128 & 0.3128 & 0.1571\\
Kev & 0.40 & 0.3988 & 0.3986 & 0.3967 & 0.3950 & 0.3988 & 0.3988 & 0.2689\\
\midrule
Base & 0.02 & 0.0200 & 0.0200 & 0.0200 & 0.0200 & 0.0200 & 0.0200 & 0.0165\\
Base & 0.05 & 0.0500 & 0.0500 & 0.0500 & 0.0500 & 0.0500 & 0.0500 & 0.0411\\
Base & 0.10 & 0.1000 & 0.1000 & 0.1000 & 0.1000 & 0.1000 & 0.1000 & 0.0813\\
Base & 0.20 & 0.2000 & 0.2009 & 0.2000 & 0.2000 & 0.2000 & 0.2000 & 0.1571\\
Base & 0.40 & 0.4400 & 0.4899 & 0.4667 & 0.4400 & 0.4400 & 0.4400 & 0.2689\\
\midrule
Posttrained & 0.02 & 0.0200 & 0.0680 & 0.0200 & 0.0200 & 0.0200 & 0.0200 & 0.0165\\
Posttrained & 0.05 & 0.0500 & 0.1146 & 0.0792 & 0.0500 & 0.0500 & 0.0500 & 0.0411\\
Posttrained & 0.10 & 0.2545 & 0.2326 & 0.1906 & 0.1724 & 0.2545 & 0.2545 & 0.0813\\
Posttrained & 0.20 & 0.4503 & 0.3388 & 0.3693 & 0.3236 & 0.4503 & 0.4503 & 0.1571\\
Posttrained & 0.40 & 0.4899 & 0.4830 & 0.4899 & 0.4828 & 0.4899 & 0.4899 & 0.2689\\
\midrule
Jev & 0.02 & 0.0200 & 0.0292 & 0.0200 & 0.0200 & 0.0200 & 0.0200 & 0.0165\\
Jev & 0.05 & 0.0502 & 0.0975 & 0.0497 & 0.0501 & 0.0501 & 0.0501 & 0.0411\\
Jev & 0.10 & 0.0965 & 0.1705 & 0.0989 & 0.0965 & 0.0964 & 0.0964 & 0.0813\\
Jev & 0.20 & 0.1853 & 0.2696 & 0.2028 & 0.1853 & 0.1844 & 0.1858 & 0.1571\\
Jev & 0.40 & 0.3109 & 0.3514 & 0.3302 & 0.3166 & 0.3093 & 0.3099 & 0.2689\\
\bottomrule\end{tabular}\end{table*}

The query costs are one for Event or Choice and two for averaging or Agreement, including the corresponding repeat controls. Always defer needs no query. The oracle is evaluator-only and is not deployable. The comparison with Random interface loss means selecting an interface uniformly at random independently of the world. It is not the better interface chosen using the evaluator posterior. Averaging and two-report equality projection are the same policy, not separate methods. No full-event graph projection or fitted reconciliation was run.

Local exact repeats are identical, so repeat controls reduce to Event. Jev repeats differ, but the preserved request order prevents a causal attribution of interface effects independent of service variation. Absolute expected loss is primary. Secondary headroom captured is $(c-\overline L)/(c-\overline L^*)$, using the same binary cohort, with its denominator recomputed in each bootstrap. It can be negative and becomes unstable near zero; normalizing does not create available improvement. Full aggregates retain denominators, flags, exact ties, and per-root helps/harms rather than clipping negative ratios.

\begin{figure*}[t]
\centering\includegraphics[width=\textwidth]{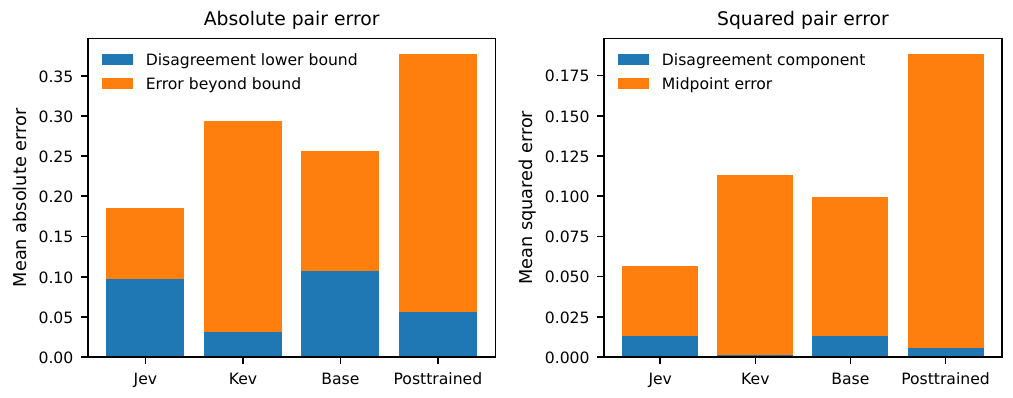}
\caption{Post-hoc Event/Choice error decomposition on all 1,000 pairs per configuration. The disagreement lower bound and the remaining component sum to the corresponding mean pair error. Ratios describe these reports and worlds; they are not causal attributions or a universal calibration bound.}
\label{fig:certificate}
\end{figure*}
\begin{figure*}[t]
\centering\includegraphics[width=\textwidth]{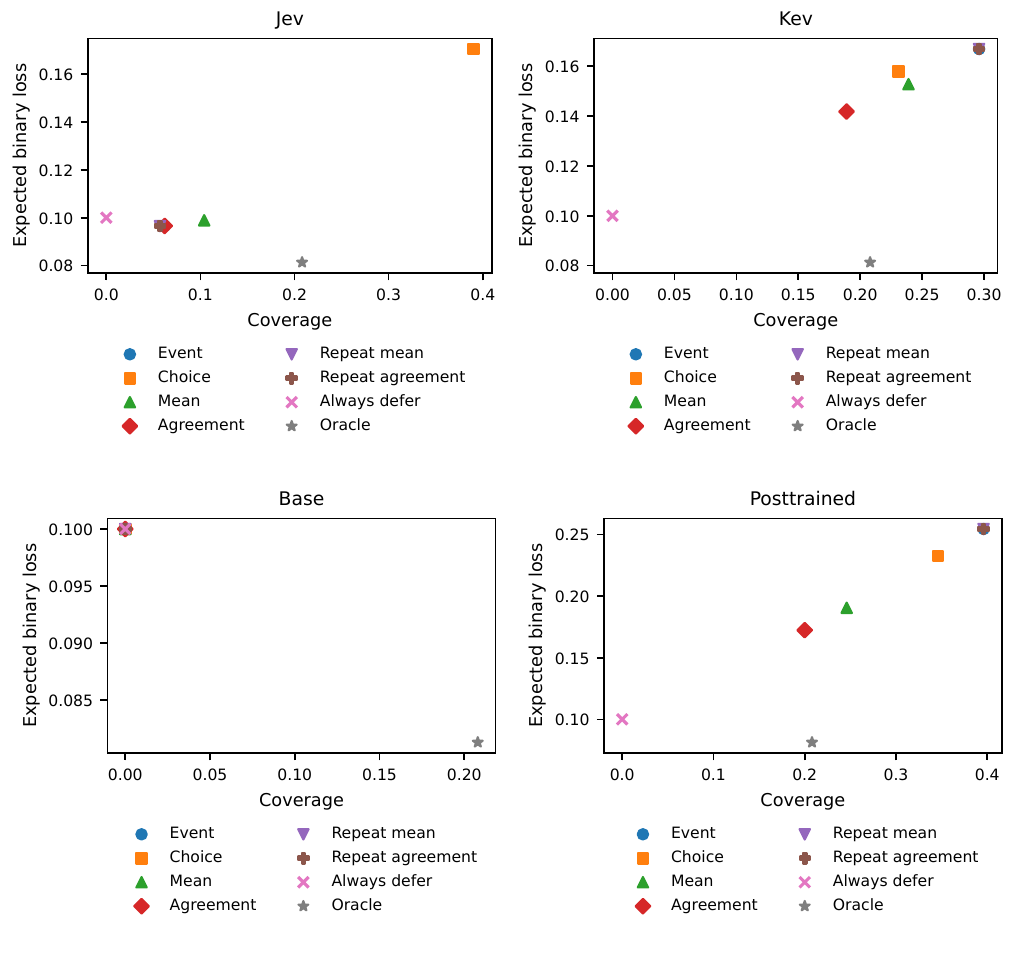}
\caption{Post-hoc fixed-policy loss and coverage at the primary defer cost $c=0.10$. Each configuration is shown separately. Always defer is a cost reference; the oracle uses evaluator truth. The policy/contrast tables supply pointwise paired intervals.}
\label{fig:policy-costs}
\end{figure*}
Figures~\ref{fig:certificate} and~\ref{fig:policy-costs} display the decomposition and primary-cost loss/coverage. Table~\ref{tab:binary-cost-grid} gives the full fixed cost grid. The secondary complement decomposition, complete policy intervals, coverage, regret, repeat controls, and helps/harms are supplied as machine-readable tables in the review artifact.

\clearpage
\raggedbottom
\section{Averaging Across Action Regions}\label{app:averaging-geometry}
This post-hoc characterization explains the fixed-policy results. It changes no collected output, policy, cost, or primary estimand. Let $0<c<1/2$, with losses $L_c(0;q)=q$, $L_c(d;q)=c$, and $L_c(1;q)=1-q$. The report-based controller selects $0$ on $[0,c]$, $d$ on $(c,1-c)$, and $1$ on $[1-c,1]$, so exact ties prefer a class over deferral.

\paragraph{Proof of Proposition 1.}
Each action region is convex. If both reports select the same action, their midpoint is in that region and $\Delta_c=0$. For two adjacent regions, the midpoint selects one of their two actions. Substituting its true loss into Equation~\ref{eq:averaging-penalty} gives half the difference between the selected and unselected losses. For opposite class actions, the Random interface has loss $\{q+(1-q)\}/2=1/2$; subtracting it from each possible midpoint action loss gives the remaining rows of Table~\ref{tab:action-regions}. These cases exhaust the action pairs. In particular, an opposite-class pair resolved to defer has penalty $c-1/2<0$. The posterior $q$ is used only to score the action, never by the policy.

\begin{table}[ht]
\centering\small
\caption{Exact per-world penalty of the mean policy versus uniformly selecting one of the two reports. The first column is an unordered action pair; rows apply when the stated midpoint action is feasible.}\label{tab:action-regions}
\begin{tabular}{lll}
\toprule
Original actions & Midpoint action & $\Delta_c$\\
\midrule
Same action & Same action & $0$\\
No, defer & No & $(q-c)/2$\\
No, defer & Defer & $(c-q)/2$\\
Defer, yes & Defer & $(c+q-1)/2$\\
Defer, yes & Yes & $(1-q-c)/2$\\
No, yes & No & $q-1/2$\\
No, yes & Defer & $c-1/2$\\
No, yes & Yes & $1/2-q$\\
\bottomrule
\end{tabular}
\end{table}

\newpage
\paragraph{An analytic high-accuracy example.}
On the original five-cost grid, the observed same-baseline penalties occur where both policies are worse than deferral. The larger grid also contains small penalties below that baseline, with intervals spanning zero. Neither empirical observation restricts what is mathematically possible for accurate reports. Fix any $c\in(0,1/2)$ and choose
\[
0<\epsilon<\min\{c,(1-2c)/3\}.
\]
Consider two equally weighted worlds. In the first, $q_1=c-\epsilon$, $p_1=c-\epsilon$, and $p'_1=c+3\epsilon$. The reports select no and defer; their midpoint $c+\epsilon$ defers. In the second, $q_2=p_2=p'_2=\epsilon$, and every policy selects no. Every report is within $4\epsilon$ of its true posterior. Averaging has population decision loss $(c+\epsilon)/2$, compared with $c/2+\epsilon/4$ for Random interface. Both are strictly better than always defer, yet averaging is worse by $\epsilon/4$. Meanwhile, averaging reduces the population expected one-coordinate Brier score by $2\epsilon^2$. Letting $\epsilon$ approach zero makes the reports arbitrarily accurate; the decision penalty also vanishes, consistent with posterior-error regret bounds. This is a constructed mathematical example, not an additional model run or evidence of performance in a high-accuracy regime.

\section{Binary Reject Loss and Elementary Scores}\label{app:scoring-bridge}

The following identity specializes standard elementary-score and mixture
representations \citep{schervish1989assessors,gneiting2007strictly,ehm2016quantiles}
to our binary reject controller and its class-before-defer tie rule.

\paragraph{Lemma.}
Let $p,q\in[0,1]$ and $0<c<1/2$. The binary controller selects no when
$p\leq c$, yes when $p\geq1-c$, and defers otherwise. Write
\[
 S_\theta^{\geq}(p;q)
 =\theta(1-q)\mathbf 1\{p\geq\theta\}
 +(1-\theta)q\mathbf 1\{p<\theta\},
\]
and define $S_\theta^{>}$ by replacing $\geq,< $ with $>,\leq$. Then
\[
 R_c(p;q)=S_c^{>}(p;q)+S_{1-c}^{\geq}(p;q).
\]
For two reports $p,p'$ with midpoint $m=(p+p')/2$, and either declared
threshold convention $h_\theta$, the corresponding elementary-score contrast
is
\[
\begin{aligned}
 &S_\theta(m;q)-\frac{S_\theta(p;q)+S_\theta(p';q)}2\\
 &\quad=(\theta-q)\left[h_\theta(m)
 -\frac{h_\theta(p)+h_\theta(p')}2\right].
\end{aligned}
\]
Thus the reject-loss contrast is the sum of the lower-strict contribution at
$c$ and upper-inclusive contribution at $1-c$, including boundary ties.
Moreover,
\[
 2\int_0^{1/2}R_c(p;q)\,dc=(p-q)^2+q(1-q).
\]
Consequently, if $C\sim\mathrm{Unif}(0,1/2)$, then
$\mathbb E_C R_C(p;q)$ equals the expected one-coordinate Brier score.
Integrating the midpoint contrast gives $-(p-p')^2/8$; the corresponding
one-coordinate Brier contrast is $-(p-p')^2/4$, and the two-coordinate
categorical convention doubles the latter.

\paragraph{Proof.}
For either threshold convention,
$S_\theta(p;q)=(1-\theta)q+(\theta-q)h_\theta(p)$. The constant cancels in
the midpoint contrast, proving the second identity. At $c$, strict thresholding
assigns equality to no; at $1-c$, inclusive thresholding assigns equality to
yes. The three controller regions then give the first identity pointwise.
Integrating the elementary score over $\theta\in(0,1)$ gives half the expected
one-coordinate Brier score, and the lower and reflected upper terms partition
that integral up to singleton boundaries. Expansion of the midpoint Brier
contrast gives the stated factors. Exact Fraction tests cover both controller
boundaries, endpoint and midpoint ties, and all 4,000 preserved Event/Choice
pairs.

This binary symmetric bridge does not imply a multiclass identity or dominance
for every proper score. Expected log-score comparisons additionally retain
their endpoint and integrability conditions.

\section{Post-Hoc Continuum and Selective Diagnostics}\label{app:accelerated-offline}

These analyses reuse the validated reports. Their specification was
recorded before computation. They use 5,000 shared, eight-stratum
root-bootstrap draws (seed 2026092408), and all intervals are pointwise
descriptive intervals. The original five-cost export was reproduced first:
1,788,000 numerical comparisons agree exactly. No prediction, validity status,
target, cost, or calibration was changed.

\subsection{Reject-cost and elementary-score curves}

The cost analysis evaluates the binary Event, Choice, mean,
Agreement, Random interface, repeat-control, always-defer, and
oracle policies at every $c=k/200$, $k=1,\ldots,99$. The machine-readable
artifact additionally gives every report, midpoint, and oracle breakpoint with
left, exact, and right values, comprising 51,366 affine open-interval segments.
For every configuration, the mean-minus-Random interface contrast takes
both signs on the presentation grid (Table~\ref{tab:v18-grid-signs}). This rules out descriptive full-domain
dominance on these reports; it does not establish population dominance or its
absence.

Figure~\ref{fig:v17-murphy-curves} is a separate Murphy diagram for the inclusive
binary elementary score over $\theta=k/200$, $k=1,\ldots,199$, with exact
breakpoint records supplied separately. Mean-versus-random elementary-score
differences likewise take both signs for every configuration. These curves
concern binary proper-score mixtures under the usual conditions. They are not
multiclass results and are not another name for the reject-cost curves.

\begin{figure*}[ht]
\centering
\includegraphics[width=\textwidth]{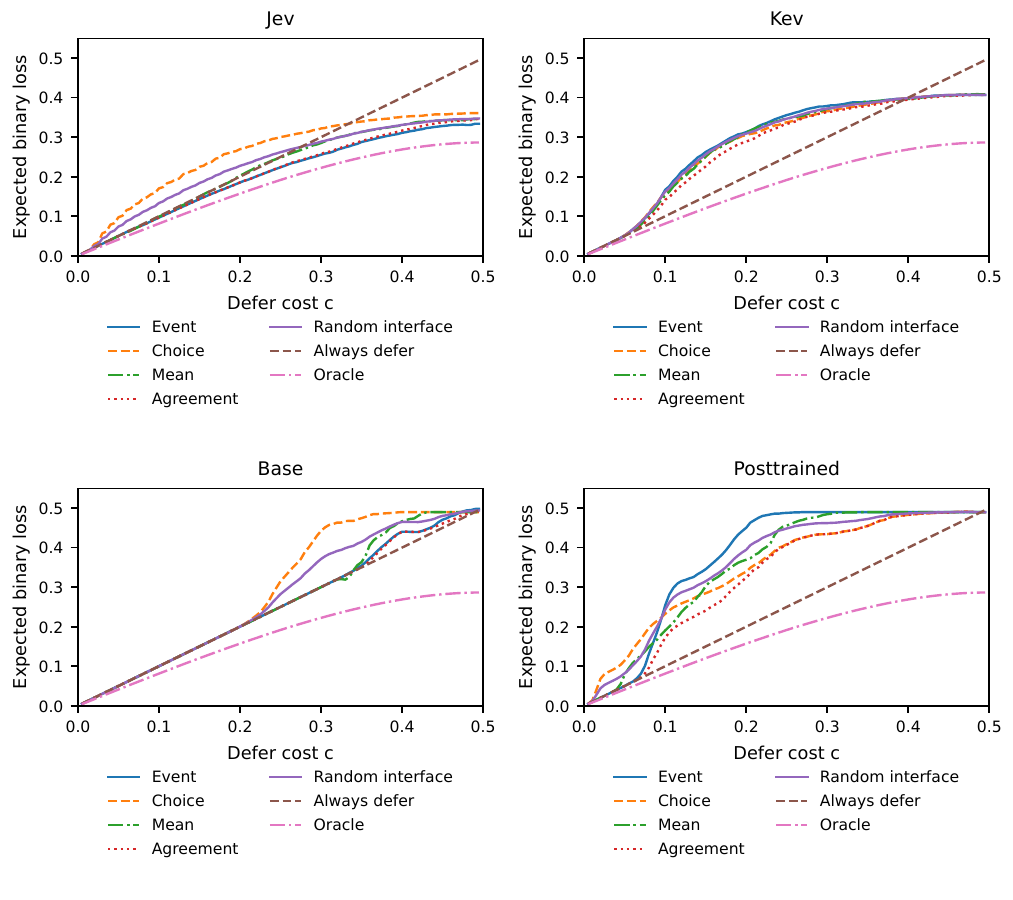}
\caption{Post-hoc expected binary reject-option loss over the symmetric defer
cost. Panels use common axes. Lines are descriptive point estimates; exact
breakpoints and pointwise bootstrap intervals are supplied in the artifact.
Lines connect the evaluated grid points and need not reproduce intervening jumps; the exact breakpoint records, rather than these lines, determine between-grid behavior.}
\label{fig:v17-reject-curves}
\end{figure*}

\begin{figure*}[ht]
\centering
\includegraphics[width=\textwidth]{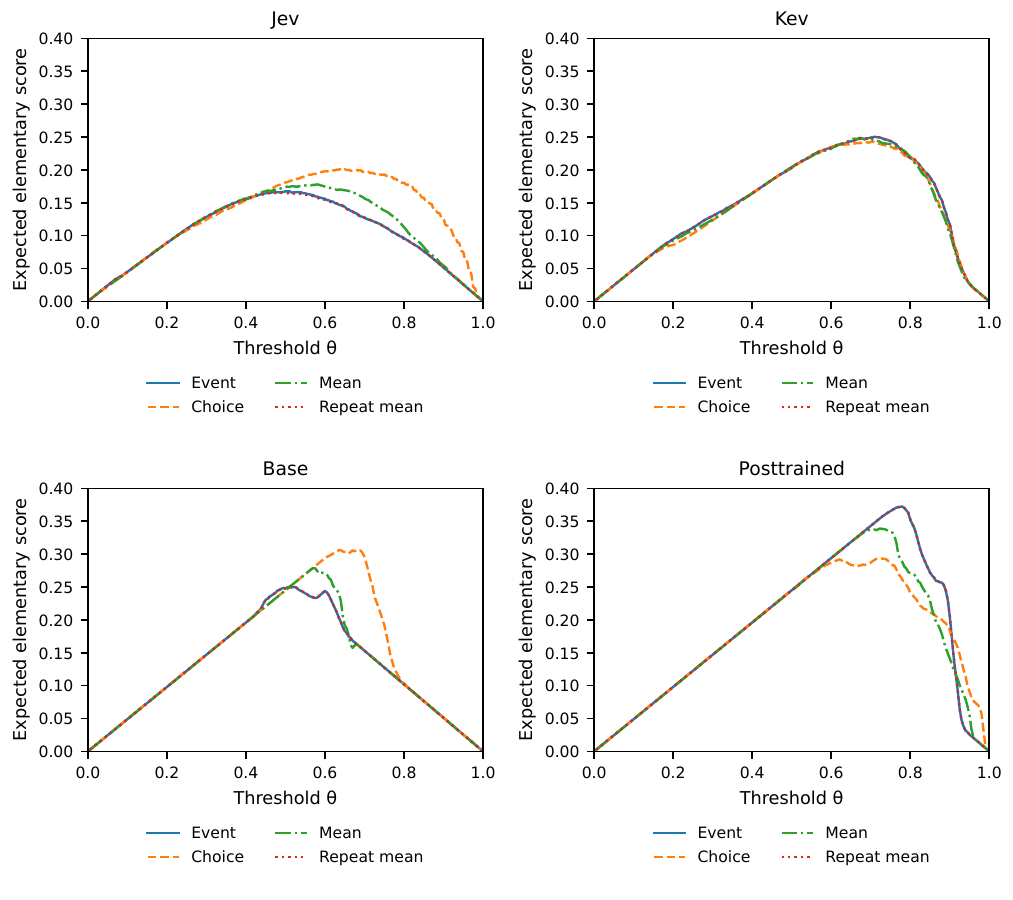}
\caption{Post-hoc expected inclusive elementary score over threshold $\theta$.
Panels use common axes. This Murphy diagram is distinct from
Figure~\ref{fig:v17-reject-curves}; it supports only the corresponding binary
proper-score-mixture interpretation. Connecting lines are a grid visualization, not the exact between-grid curve.}
\label{fig:v17-murphy-curves}
\end{figure*}

\begin{table}[ht]
\centering\small\setlength{\tabcolsep}{4pt}
\caption{Signs of mean-minus-Random interface expected loss on the post-hoc 99-cost grid. Counts concern grid locations, not independent trials or shares of the cost continuum. ``Both below'' counts positive contrasts for which both policies beat always defer.}\label{tab:v18-grid-signs}
\begin{tabular}{lrrrr}
\toprule
Configuration & Lower & Equal & Higher & Both below\\
\midrule
Jev & 82 & 1 & 16 & 16\\
Kev & 62 & 5 & 32 & 7\\
Base & 47 & 36 & 16 & 0\\
Posttrained & 47 & 1 & 51 & 2\\
\bottomrule
\end{tabular}
\end{table}

There are 25 grid locations where averaging has higher point-estimated loss while both policies beat deferral: 16 for Jev, seven for Kev, and two for posttrained. Every associated pointwise 95\% contrast interval includes zero. For scale, at $c=0.42$, Jev's mean and Random interface losses are 0.33856 and 0.33629; their difference is 0.00227 [$-0.00159$, 0.00621], below always-defer loss 0.42 for both policies. This cost is a post-hoc illustration, not a prespecified endpoint. The full grid is retained; no multiplicity-adjusted or population-wide harm claim follows. The five-cost observation therefore does not extend to the denser grid.

\subsection{Signed complement mass and paired strata}

The signed complement calculation reproduces the earlier absolute residuals.
The mean sums $p(A)+p(\neg A)$ are 1.06414, 1.50547, 1.11033, and 1.80003 for
Jev, Kev, base, and posttrained. The fraction exceeding 1.5 is
\VxJevComplementAboveOnePointFive, \VxKevComplementAboveOnePointFive,
\VxBaseComplementAboveOnePointFive, and
\VxPostComplementAboveOnePointFive, respectively. None is below 0.5. These
distributions describe excess affirmative mass but do not identify its cause.

On 990 roots with valid canonical reports from both Kev and Jev, the equally
weighted mean of the eight paired stratum differences (Kev minus Jev TV) is
\VxKevMinusJevPairedTV{} (pointwise 95\% interval
[\VxKevMinusJevPairedTVLower, \VxKevMinusJevPairedTVUpper]). Four stratum
means are positive and four negative. This is a paired-support estimand, distinct
from the earlier marginal 4--4 ordering.

The primary interval in Section~\ref{sec:results} remains the main reference. The interval above resamples full stratum root slots with valid-count renormalization, rather than the specified paired-valid support. Direct paired-support resampling gives $-0.033451$ [$-0.045113$, $-0.022154$], with the same qualitative conclusion. Both procedures and the specification deviation are documented in the artifact.

\subsection{Disagreement as a selective signal}

The post-hoc selector retains the lowest Event/Choice disagreement and uses the
mean report's binary class on retained roots. At exactly matched coverage it is
compared with decreasing mean-report confidence, uniform random retention, and
an evaluator-only ordering. Ties at the retention boundary receive equal
fractional weight. Table~\ref{tab:v18-selector-midpoint} shows every configuration at the specified midpoint coverage. Disagreement has lower point-estimated loss than random retention for Jev and posttrained, but lower loss than confidence retention only for Jev. All three deployable/random strategies exceed always-defer loss 0.10 at this coverage. The full coverage grid, not a selected favorable coverage, remains in the artifact.

\begin{table}[ht]
\centering\small\setlength{\tabcolsep}{4pt}
\caption{Post-hoc binary pipeline loss at 50\% retained coverage and $c=0.10$. Every strategy uses the same mean-report class predictor on retained worlds; unretained worlds defer. Entries are point estimates; the artifact supplies paired uncertainty summaries.}\label{tab:v18-selector-midpoint}
\begin{tabular}{lrrr}
\toprule
Configuration & Disagreement & Confidence & Random\\
\midrule
Jev & 0.17311 & 0.17943 & 0.22314\\
Kev & 0.25827 & 0.23729 & 0.25416\\
Base & 0.29737 & 0.29122 & 0.29496\\
Posttrained & 0.29110 & 0.27650 & 0.29496\\
\bottomrule
\end{tabular}
\end{table}

\begin{figure*}[ht]
\centering
\includegraphics[width=\textwidth]{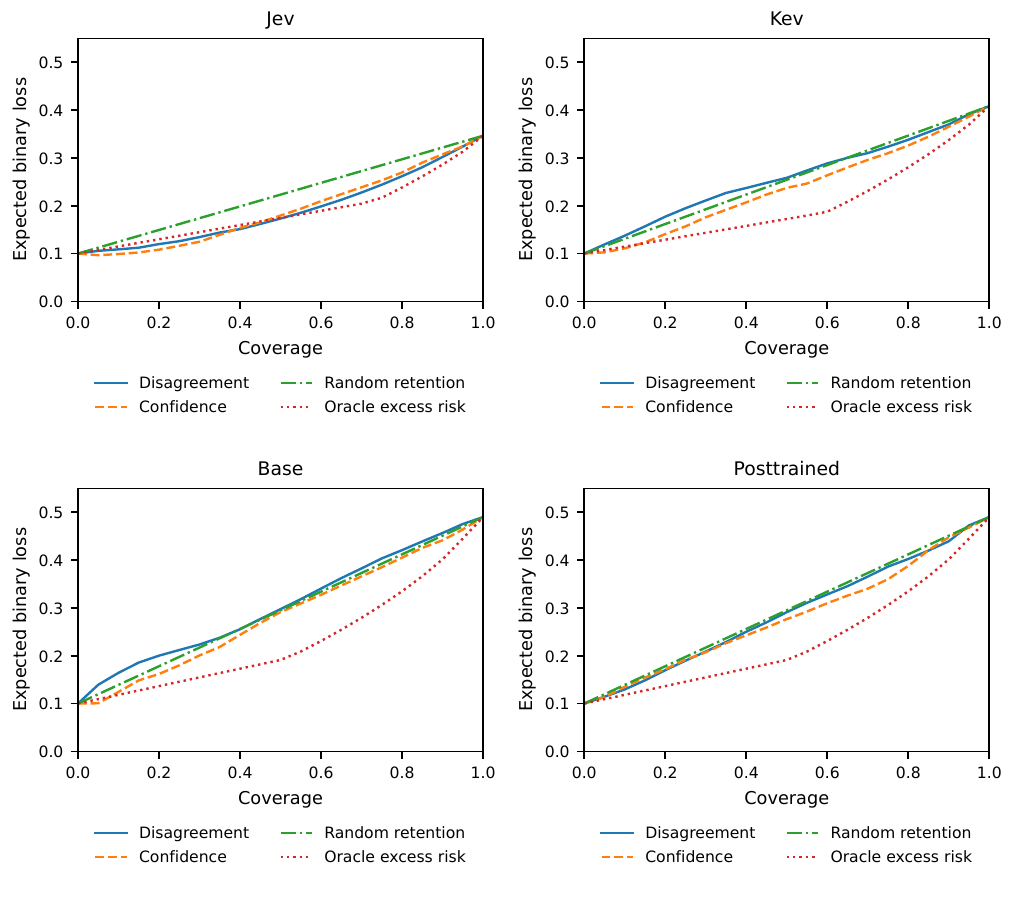}
\caption{Post-hoc binary pipeline loss at $c=0.10$ versus retained coverage.
Panels use common axes. Disagreement, confidence, and random retention share the
same fixed class predictor and candidate population; the oracle-excess ordering
uses evaluator truth and is nondeployable. It ranks class excess risk and is not an optimal envelope for pipeline loss. At full coverage this is a binary
class predictor without the original reject option.}
\label{fig:v17-selective}
\end{figure*}

For Base and posttrained, the mean-report class predictor answers yes on all 1,000 roots. Their identical random-retention loss of 0.29496 is therefore expected: random retention applies the same constant predictor on the same targets. Confidence and disagreement can still select different subsets of worlds, so their selective losses need not agree. This concerns the forced class predictor, not the reject-option controller.

The evaluator-only ordering ranks \emph{class excess risk}, not absolute class risk. It therefore need not minimize pipeline loss at fixed coverage: roots with a Bayes-optimal class can still have different irreducible error. It is a diagnostic comparator, not a lower bound on the displayed loss. A minimum-pipeline-loss ordering would instead retain the lowest exact expected loss of the fixed class predictor. No such additional ordering replaces the frozen comparator here.

The tie-aware AUROC for using high disagreement to identify strictly positive
mean-controller regret at $c=0.10$ is \VxJevDisagreementAUC,
\VxKevDisagreementAUC, \VxBaseDisagreementAUC, and
\VxPostDisagreementAUC{} for Jev, Kev, base, and posttrained. This diagnostic concerns regret of a reject-option controller, not correctness of its class prediction. Correct deferral can coexist with a poor forced class prediction, so the below-0.5 values do not establish a general classifier-error ranking or a useful deployable detector. The implementation labels regret above
$10^{-15}$ as positive. An independent exact-rational check agrees with those
labels on all 4,000 current roots; this does not establish equivalence for future
data. In a separate supplemental experiment,
the selector can only remove existing class actions; retaining all active roots
exactly reproduces the original mean-policy loss.

The sampled-outcome calibration summaries retain their prespecified status;
the exact-posterior summaries retain their post-hoc status. No recalibration
was fitted.

\clearpage
\section{Reasoning-Budget Diagnostic and Exploratory Scores}\label{app:reasoning-budget}
The original zero-valid reasoning results in Appendix~\ref{app:prospective-stress} are unchanged. A later diagnostic tested completion with a larger token budget. Its outputs were subsequently scored in a separate exploratory analysis; the metrics and reporting populations were fixed after collection and before score inspection. This analysis neither satisfies the original feasibility criterion nor supplies a new confirmatory comparison.

\paragraph{Original cap and later configuration.}
At a 4,096-token development limit, six of 32 thinking traces completed in 2,779--4,042 tokens and 26 reached the limit; all 32 non-thinking reports were valid. A runtime projection selected the original 160-token main cap, below every completed development trace. The artifact supplies the ledger and cap-selection rule; the original raw development traces remain unavailable.

The later diagnostic planned 160 requests on 16 fresh roots: canonical, Event, Choice, complement, and repeat, each in thinking and non-thinking modes. It used \mbox{Qwen3.5-4B} revision \texttt{851bf6e806efd8d0a36b00ddf55e13ccb7b8cd0a}, BF16, and batch size one. The total continuation allowance was 32,768 tokens, with 32,763 available for reasoning. A valid thinking report required a model-generated terminator before the answer-ID likelihood readout. No synthetic terminator, retry, repaired output, or evaluator target was used in collection. The scored probabilities are normalized answer-token likelihoods, not numerical forecasts emitted directly by the model.

\paragraph{Support and analysis populations.}
Collection spanned three runtime stages (Table~\ref{tab:reasoning-diagnostic-accounting}). The final stage assigned whole mode pairs to four A100 replicas with matched optimized kernels. Previously acquired requests, including interruptions and cap failures, were excluded from further acquisition. Across the ledger, 159 planned identities were acquired, 157 have terminal records, two were interrupted, and one was never acquired. The original criterion required all 160 terminal records and minimum valid counts by request type. It remains unmet: canonical and binary-Choice thinking each have 13 valid reports across stages, below the required 15. No larger study followed.

\begin{table}[ht]
\centering\small
\caption{Operational diagnostic counts by runtime stage and mode. Valid and capped requests have terminal records; interrupted requests do not. Counts describe the acquired requests in each stage, not estimates of accuracy or a pooled completion rate. One further planned non-thinking complement request was never acquired.}
\label{tab:reasoning-diagnostic-accounting}
\begin{tabular}{llrrrr}
\toprule
Runtime stage & Mode & Acquired & Valid & Capped & Interrupted\\
\midrule
Original & Thinking & 6 & 3 & 2 & 1\\
Original & Non-thinking & 6 & 6 & 0 & 0\\
Optimized single replica & Thinking & 5 & 4 & 0 & 1\\
Optimized single replica & Non-thinking & 4 & 4 & 0 & 0\\
Four optimized replicas & Thinking & 69 & 64 & 5 & 0\\
Four optimized replicas & Non-thinking & 69 & 69 & 0 & 0\\
\bottomrule
\end{tabular}
\end{table}

Scores use exact posterior targets reconstructed from the 16 world definitions and their event maps. Mode comparisons require the same root, request type, runtime stage, and two valid reports. Each reported mean weights its eligible roots equally. Failure-fallback loss is reported separately on each stage's intended identities, assigning missing or invalid reports to defer without imputing probabilities. Runtime stages are kept separate because their inputs and kernels are not established to be bitwise equivalent.

\paragraph{Matched posterior scores and decisions.}
In the final stage, 64 of 69 thinking reports and all 69 non-thinking reports are valid. Table~\ref{tab:reasoning-exploratory} gives all five request-type comparisons. Thinking has lower mean TV and expected Brier on the matched-valid supports. At $c=0.10$, however, every valid thinking report selects a class, while only three of its 64 matched controls do so. On the 12 matched canonical roots, thinking TV is 0.3845 versus 0.4633 for the control, but expected loss is 0.3902 versus 0.1000. Reduced posterior error therefore coexists with actions whose true error exceeds the deferral cost.

\begin{table}[ht]
\centering\small
\caption{Exploratory matched thinking-minus-non-thinking differences in the final runtime stage. Each row uses the same $n$ roots with valid reports in both modes. Negative values favor thinking. TV and expected Brier concern the stated request partition; binary Brier includes both coordinates. These are descriptive means from at most 16 worlds, with dependent request types and no confidence intervals. All declared costs, earlier stages, root IDs, and failure-fallback scores are supplied in the artifact.}
\label{tab:reasoning-exploratory}
\begin{tabular}{lrrrrr}
\toprule
Request & $n$ & $\Delta$TV & $\Delta$Brier & $\Delta L_{0.10}$ & $\Delta L_{0.40}$\\
\midrule
Canonical & 12 & -0.0789 & -0.0581 & +0.2902 & -0.1969\\
Event & 15 & -0.1331 & -0.1517 & +0.1068 & -0.3008\\
Choice & 12 & -0.2735 & -0.4217 & +0.0529 & -0.3893\\
Complement & 13 & -0.0423 & -0.0287 & +0.1562 & -0.1480\\
Repeat & 12 & -0.0968 & -0.1138 & +0.1313 & -0.2911\\
\bottomrule
\end{tabular}
\end{table}

The decision ordering depends on cost: at $c=0.40$, all five matched loss contrasts favor thinking. The artifact reports every value on the original grid $c\in\{0.02,0.05,0.10,0.20,0.40\}$ and the separately specified secondary grid that also includes 0.30. These mode comparisons are distinct from the Mean-versus-Random-interface averaging experiment in Section~7. Neither implies that reasoning generally harms or improves decisions.

\paragraph{Other stages, relations, and limitations.}
The first stage has one valid mode pair each for canonical, Choice, and repeat; repeat worsens in TV and Brier. The optimized single-replica stage has one valid complement pair, which worsens, and three repeat pairs, which improve. These sparse cells and all undefined cells remain in the artifact.

In the final stage, marginal Event/Choice absolute residuals are 0.010634 for thinking on 12 complete pairs and 0.218515 for non-thinking on 14. Complement residuals are 0.250952 on 12 thinking pairs and 0.409232 on 13 non-thinking pairs. These mode means use different supports and are not paired mode-effect estimates. Exact supports and signed residuals are provided separately.

The diagnostic gives conditional descriptive evidence from 16 worlds, not five independent replications or 138 independent cases. Completion-dependent missingness, runtime changes, and small supports limit interpretation; no confidence intervals or significance tests were computed. The remaining canonical posterior error does not establish a high-accuracy regime. The artifact preserves every intended identity and supplies both the original completion-only records and the later exploratory scores, with independent CPU verification.

\end{document}